\documentclass[10pt,twocolumn,letterpaper]{article}

\usepackage[accsupp]{axessibility}
\usepackage{cvpr}              
\usepackage{graphicx}
\usepackage{amsmath}
\usepackage{amsthm}
\usepackage{booktabs}
\usepackage{algorithm}
\usepackage{algorithmic}
\usepackage{epstopdf}
\usepackage{makecell}
\usepackage{color}
\usepackage{alltt}
\usepackage{multirow}
\usepackage{float}
\usepackage{wrapfig}
\usepackage{adjustbox}
\usepackage{makecell}
\usepackage[dvipsnames]{colortbl}
\usepackage{subcaption}
\usepackage{cancel}
\usepackage{amssymb}
\PassOptionsToPackage{dvipsnames}{xcolor}
\usepackage[dvipsnames]{xcolor}

\definecolor{cvprblue}{rgb}{0.21,0.49,0.74}
\usepackage[pagebackref,breaklinks,colorlinks,citecolor=cvprblue,backref=page]{hyperref}

\def\paperID{1917} 
\def\confName{CVPR}
\def\confYear{2024}
\definecolor{olive}{rgb}{0.5, 0.5, 0.0}
\definecolor{maroon}{rgb}{0.69, 0.19, 0.38}
\definecolor{celestialblue}{rgb}{0.29, 0.59, 0.82}
\definecolor{darkgreen}{rgb}{0.0, 0.6, 0.0}
\definecolor{grey}{rgb}{0.5,0.5,0.5}
\definecolor{darkblue}{rgb}{0.19, 0.19, 0.62}
\definecolor{silver}{rgb}{0.7,0.7,0.7}
\definecolor{darkcyan}{rgb}{0.0, 0.55, 0.55}

\def\clap#1{\hbox to 0pt{\hss #1\hss}}%

\newcommand\undefcolumntype[1]{\expandafter\let\csname NC@find@#1\endcsname\relax}

\definecolor{C0}{rgb}{0.121569, 0.466667, 0.705882}
\definecolor{C1}{rgb}{1.000000, 0.498039, 0.054902}
\definecolor{C2}{rgb}{0.172549, 0.627451, 0.172549}
\definecolor{C3}{rgb}{0.839216, 0.152941, 0.156863}
\definecolor{C4}{rgb}{0.580392, 0.403922, 0.741176}
\definecolor{C5}{rgb}{0.549020, 0.337255, 0.294118}
\definecolor{C6}{rgb}{0.890196, 0.466667, 0.760784}
\definecolor{C7}{rgb}{0.498039, 0.498039, 0.498039}
\definecolor{C8}{rgb}{0.737255, 0.741176, 0.133333}
\definecolor{C9}{rgb}{0.090196, 0.745098, 0.811765}

\newcommand{\baseline}[1]{\textit{\textcolor{gray}{#1}}}
 
\newlength\savewidth\newcommand\shline{\noalign{\global\savewidth\arrayrulewidth
  \global\arrayrulewidth 1pt}\hline\noalign{\global\arrayrulewidth\savewidth}}
\definecolor{Blue}{RGB}{0, 0, 255}
\definecolor{Aquamarine}{RGB}{127, 255, 212}
\definecolor{Sepia}{RGB}{112, 66, 20}
\definecolor{BrickRed}{RGB}{203, 65, 84}
\newcommand{\blue}{\textcolor[rgb]{0.173, 0.451, 0.824}}
\colorlet{my-red}{BrickRed!90!Sepia}
\colorlet{my-blue}{Aquamarine!30!Blue}

\definecolor{ForestGreen}{RGB}{34,139,34}
\definecolor{Black}{RGB}{0,0,0}

\newcommand{\hll}[1]{\textcolor{black}{#1}}

\title{Generalized Large-Scale Data Condensation via Various Backbone and Statistical Matching}

\author{Shitong Shao$^{1,3}$ \qquad Zeyuan Yin$^{1}$ \qquad Muxin Zhou$^{1}$ \qquad Xindong Zhang$^{2,3}$ \qquad Zhiqiang Shen$^{1,*}$ \\
  $^1$Mohamed bin Zayed University of AI $^2$Hong Kong Polytechnic University $^3$OPPO Research \\
  {\tt\small 1090784053sst@gmail.com, \{zeyuan.yin,muxin.zhou,zhiqiang.shen\}@mbzuai.ac.ae} \\
   {\tt\small 17901410r@connect.polyu.hk, $*$:Corresponding author}}
\begin{document}
\twocolumn[{%
\renewcommand\twocolumn[1][]{#1}%
\maketitle\begin{center}\centering\includegraphics[width=1.0\linewidth,height=0.26\textheight]{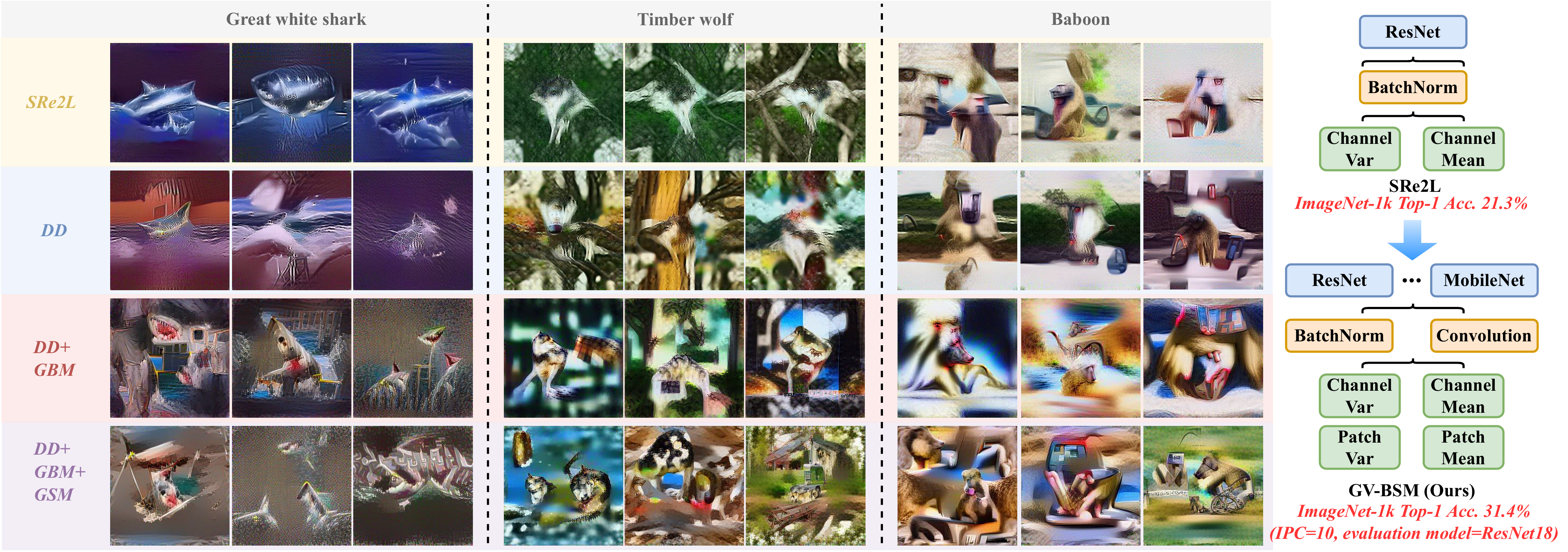}
\captionof{figure}{\textbf{Left:} Our proposed \textbf{G-VBSM} consists of three novel and effective modules, named  DD, GSM and GBM. The richness and quality of information in the synthetic data have been significantly enhanced compared with the \baseline{baseline} SRe2L through the sequential merging of DD, GBM, and GSM. \textbf{Right:} G-VBSM prioritizes ``generalized matching'' to ensure consistency between distilled and complete datasets across various backbones, layers, and statistics, and achieves the highest accuracy 31.4\% on ImageNet-1k under IPC 10.}
\label{fig:first_presentation}
\end{center}%
}]

\begin{abstract}
\hll{The lightweight ``local-match-global" matching introduced by SRe2L successfully creates a distilled dataset with comprehensive information on the full 224$\times$224 ImageNet-1k. However, this one-sided approach is limited to a particular backbone, layer, and statistics, which limits the improvement of the generalization of a distilled dataset. We suggest that sufficient and various ``local-match-global'' matching are more precise and effective than a single one and have the ability to create a distilled dataset with richer information and better generalization ability.} We call this perspective ``generalized matching'' and propose \textbf{G}eneralized \textbf{V}arious \textbf{B}ackbone and \textbf{S}tatistical \textbf{M}atching (\textbf{G-VBSM}) in this work, which aims to create a synthetic dataset with densities, ensuring consistency with the complete dataset across various backbones, layers, and statistics. As experimentally demonstrated, G-VBSM is the first algorithm to obtain strong performance across both small-scale and large-scale datasets. Specifically, G-VBSM achieves performances of 38.7\% on CIFAR-100, 47.6\% on Tiny-ImageNet, and 31.4\% on the full 224×224 ImageNet-1k, respectively\footnote{Settings: CIFAR-100 with 128-width ConvNet under 10 images per class (IPC), Tiny-ImageNet with ResNet18 under 50 IPC, and ImageNet-1k with ResNet18 under 10 IPC.}. These results surpass all SOTA methods by margins of 3.9\%, 6.5\%, and 10.1\%, respectively.
\end{abstract}
\vspace{-10pt}    
\vspace{-8pt}
\section{Introduction}
\label{sec:intro}
With the development of deep learning, the number of model parameters and the quantity of training data have become increasingly large~\cite{Zagoruyko2016WRN,kdsurvey}. Researchers have tried to minimize the training overhead while preventing a decline in the generalization ability. Data condensation (DC), also known as Dataset distillation, first introduced by Wang \textit{et al.}~\cite{dd_begin}, aims to alleviate the training burden by synthesizing a small yet informative distilled dataset derived from the complete training dataset, while ensuring that the behavior of the distilled dataset on the target task remains consistent with that of the complete dataset. The extremely compressed distilled dataset contains sufficiently valuable information and have the potential for fast model training, and have been become a popular choice for different downstream application, like federated learning~\cite{dd_federated_learning_1,dd_continual_learning_2}, continual learning~\cite{dd_continual_learning_1,dd_continual_learning_2,dd_continual_learning_3}, neural architecture search~\cite{dd_nas_1,dd_dist_matching,dd_gradient_matching} and 3D point clouds~\cite{Wang3dpoint}.
\begin{figure*}[!t]
\begin{center}
\includegraphics[width=0.9\linewidth,trim={0cm 0.7cm 0cm 0cm},clip]{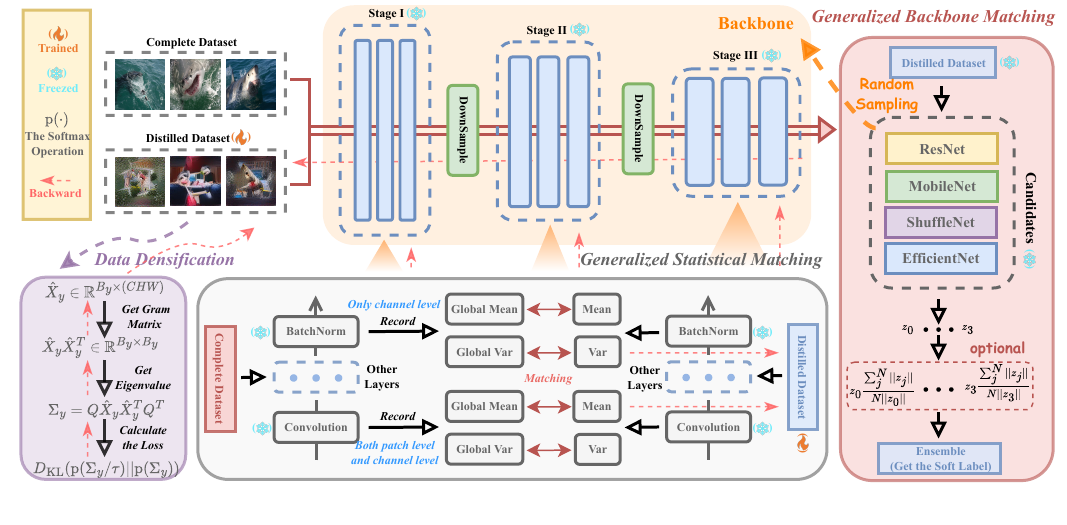}
\vspace{-10pt}
\end{center}
   \caption{The overview of G-VBSM on the full 224$\times$224 ImageNet-1k, which ensures the consistency between the distilled and the complete datasets across various backbones, layers and statistics via ``generalized matching''.}
\label{fig:total_framework}
\vspace{-10pt}
\end{figure*}

A persistent problem that researchers have been working to solve~\cite{dd_efficient_parameterization,dd_gradient_matching,dd_dream,dd_tesla} in DC is the substantial data synthesis overhead~\cite{dd_survey,dd_comprehensive_review}, which hinders its application in real-world large-scale datasets (\textit{e.g.}, ImageNet-1k) training. Typical performance matching~\cite{dd_begin,dd_FRePo,dd_kip} and trajectory matching~\cite{dd_mtt,dd_minimizing_acc_traj_error} unroll recursive computation graphs, requiring substantial GPU memory and resulting in prohibitive training costs. Zhao \textit{et al.}~\cite{dd_gradient_matching} proposed gradient matching to address this, synthesizing distilled datasets by matching gradients from synthetic and real data in a single step. However, gradient computation and matching remain time-consuming~\cite{dd_efficient_parameterization}, leading to the proposal of distribution matching~\cite{dd_dist_matching}. This method and its variants~\cite{dd_CAFE} employ a network-based feature extractor to embed both synthetic and real data into a high-dimensional Hilbert space, then perform distribution matching. The training load for this direct, single-step process stems only from one gradient update of the synthetic data and the feature extractor~\cite{dd_comprehensive_review}. Unfortunately, all of the above mentioned improved methods still have extremely large training overheads on the full 224$\times$224 ImageNet-1k.


Recently, SRe2L~\cite{dd_sre2l} accomplished data condensation for the first time on the full 224$\times$224 ImageNet-1k~\cite{ILSVRC15}, achieving Top-1 validation accuracy 21.3\% with ResNet18 under IPC 10. This method outperformed the latest state-of-the-art TESLA~\cite{dd_tesla}, which conducted on a low-resolution version of ImageNet-1k, by being 16$\times$ faster and improved performance by a margin of 13.6\%. SRe2L is inspired by DeepInversion~\cite{deepinversion} and aims to match statistics in BatchNorm generated from synthetic and real data. We reevaluate the success of SRe2L through the lightweight ``local-match-global'' essentially. The ``local-match-global'' refers to utilizing more comprehensive information (\textit{e.g.}, statistics in BatchNorm), generated from the model using the complete dataset (global), to guide the parameter update of the distilled dataset (local).

\hll{However, such lightweight and effective matching in SRe2L is singular, depending on the particular layer (\textit{i.e.}, BatchNorm), model (\textit{i.e.}, ResNet18), and statistics (\textit{i.e.}, channel mean/variance). Intuitively, sufficient ``local-match-global'' matching can result in more accurate and rational supervision than a single one, further enhancing the generalization of the distilled dataset. We call this perspective ``generalized matching'' and suggest that the distilled dataset is likely to perform consistent with the complete dataset on the evaluation model, provided that there is sufficient variety in backbones, layers, and statistics used for matching.}

Inspired by \hll{this}, we propose \textbf{G}eneralized \textbf{V}arious \textbf{B}ackbone and \textbf{S}tatistical \textbf{M}atching (\textbf{G-VBSM}) to \hll{fulfill ``generalized matching''}. G-VBSM comprises three integral and effective parts named data densification (DD), generalized statistical matching (GSM), and generalized backbone matching (GBM). \hll{DD is utilized to ensure that the images within each class are linearly independent, thereby enhancing the (intra-class) diversity of the distilled dataset. This ultimately guarantees that ``generalized matching'' preserves the rich and diverse information within the synthetic data. GBM and GSM are designed to implement ``generalized matching''. Among them, GSM traverses the complete dataset without computing and updating the gradient, to record the statistics of Convolution at both the patch and channel levels. These statistics are subsequently utilized for matching during the data synthesis phase, in conjunction with the channel-level statistics in BatchNorm. Furthermore, GBM aims to ensure consistency between distilled and complete datasets across various backbones, enhancing matching sufficiency and leading to strong generalization in the evaluation phase.} In particular, G-VBSM also ensures the efficiency of dataset condensation through a series of strategies, as mentioned in Sec.~\ref{sec:method}.

Extensive experiments on CIFAR-10, CIFAR-100, Tiny-ImageNet, and the full 224$\times$224 ImageNet-1k, demonstrating that G-VBSM is the first algorithm that performs well on both small-scale and large-scale datasets. Specifically, we not only verify through ablation studies that GSM, GBM and DD are consistently reliable, but also achieve the highest 38.7\%, 47.6\% and 31.4\% on CIFAR-100 (128-width ConvNet), Tiny-ImageNet (ResNet18), and the full 224$\times$224 ImageNet-1k (ResNet18) under images per class (IPC) 10, 50 and 10, respectively, which outperforms all previous state-of-the-art (SOTA) methods by 3.9\%, 6.5\% and 10.1\%, respectively.
\section{Background}
\label{sec:background}
Dataset condensation (DC) represents a data synthesis procedure that aims to compress a complete, large dataset $\mathcal{T} = \{(X_i,y_i)\}_{i=1}^{|\mathcal{T}|}$ into a smaller, distilled dataset $\mathcal{S} = \{(\tilde{X}_i,\tilde{y}_i)\}_{i=1}^{|\mathcal{S}|}$, subject to ${|\mathcal{S}|} \ll {|\mathcal{T}|}$, while ensuring that an arbitrary evaluation model $f_\textrm{eval}(\cdot)$ trained on $\mathcal{S}$ yields similar performance to one trained on $\mathcal{T}$. Classical data distillation algorithms invariably require the candidate model $f_\textrm{cand}(\cdot)$ to execute one or more steps on 
$\mathcal{S}$ to update its parameter $\theta_\textrm{cand}$, subsequently achieving matching in terms of performance~\cite{dd_begin,dd_kip}, gradient~\cite{dd_gradient_matching}, trajectory~\cite{dd_mtt,dd_tesla}, or distribution~\cite{dd_dist_matching,dd_CAFE}. The process $\theta_\textrm{cand} - \alpha \nabla_{\theta_\textrm{cand}}\ell (f_\textrm{cand}(\tilde{X}),\tilde{y})$, where $\ell(\cdot,\cdot)$ and $(\tilde{X},\tilde{y})$ represent the loss function and a batch sampled from $\mathcal{S}$, respectively, is notably time-consuming. Consequently, even the relatively swiftest distribution matching~\cite{dd_CAFE,dd_datadam} is slower than the recent proposed SRe2L~\cite{dd_sre2l}. In fact, SRe2L is the only workable way to achieve DC on the full 224$\times$224 ImageNet-1k, as it requires updating the parameters of the synthetic data only once per iteration.

SRe2L~\cite{dd_sre2l} encompasses three incremental subprocesses: Squeeze, Recover, and Relabel. Squeeze is designed to train $f_\textrm{cand}(\cdot)$ containing BatchNorm in a standard manner, aiming to record the global channel mean ${\textbf{\textrm{BN}}_l^\textrm{CM}}$ and channel variance ${\textbf{\textrm{BN}}_l^\textrm{CV}}$ ($l$ refers to the index of the $l$-th layer) via exponential moving average (EMA), extracted from $\mathcal{T}$, for subsequent matching in Recover. In Recover after that, given the channel mean $\mu_l(\tilde{X})$ and channel variance $\sigma^2_l(\tilde{X})$ in $l$-th BatchNorm obtained from $\mathcal{S}$, the statistical matching loss function can be formulated as
\begin{equation}
\footnotesize
\begin{aligned}
& \mathcal{L}_\textrm{BN}(\tilde{X}) = \sum_l \left|\left|\mu_l(\tilde{X}) - {\textbf{\textrm{BN}}_l^\textrm{CM}}\right|\right|_2 + \left|\left|\sigma^2_l(\tilde{X}) - {\textbf{\textrm{BN}}_l^\textrm{CV}}\right|\right|_2.\\
\end{aligned}
\label{eq:sre2l_bn_function}
\end{equation}
Based on this, we can give the entire optimization objective in Recover as
\begin{equation}
\footnotesize
\begin{aligned}
& \operatorname*{arg\,min}_{\tilde{X}} \mathcal{L}_\textrm{BN}(\tilde{X}) + \ell (f_\textrm{cand}(\tilde{X}),y),\\
\end{aligned}
\label{eq:sre2l_entire_function}
\end{equation}
where $y$ denotes the ground truth label. Moreover, SRe2L assigns soft labels $\tilde{y}$ to the synthetic data, utilizing the logit-based distillation~\cite{DIST,vanillakd} to improve the generalization ability of the distilled dataset. This can be denoted as
\begin{equation}
\small
\begin{aligned}
& \tilde{y} = \textrm{softmax}(f_\textrm{cand}(\tilde{X})/\tau),\\
\end{aligned}
\label{eq:sre2l_relabel_function}
\end{equation}
where $\tau$ denotes the temperature to regulate the smoothness of the soft labels, thereby enhancing the distilled dataset's potential for generalization to unseen evaluation models. The generated soft label can be stored on disk using FKD~\cite{shen2022fast} so as not to defeat the purpose of DC. A crucial point of SRe2L is that BatchNorm calculates the statistics of the entire dataset using EMA, thereby providing a comprehensive and representative matching for the distilled dataset. Encouraged by this, \hll{our research focuses on applying sufficient ``local-match-global'' matching to achieve ``generalized matching''}.

\section{Method}
\label{sec:method}

The comprehensive and detailed framework of our proposed \textbf{G}eneralized \textbf{V}arious \textbf{B}ackbone and \textbf{S}tatistical \textbf{M}atching (\textbf{G-VBSM}) is illustrated in Fig.~\ref{fig:total_framework}. \hll{In essence, G-VBSM employs the lightweight regularization strategy data densification (DD) to ensure both the diversity and density of the distilled dataset, ensuring that the potential of ``generalized matching'' can be fully exploited. Moreover, generalized backbone matching (GBM) and generalized statistical matching (GSM) are utilized to achieve ``generalized matching'' by performing ``local-match-global'' matching across various backbones, layers, and statistics.} In particular, the efficiency and effectiveness of DD, GBM, GSM, SRe2L, and TESLA are illustrated in Fig.~\ref{fig:efficiency_vs_effectiveness}.

\subsection{Data Densification}
\label{sec:method_dd}
\begin{figure}[!h]
\begin{center}
\includegraphics[width=0.95\linewidth,trim={0cm 0.2cm 0cm 0cm},clip]{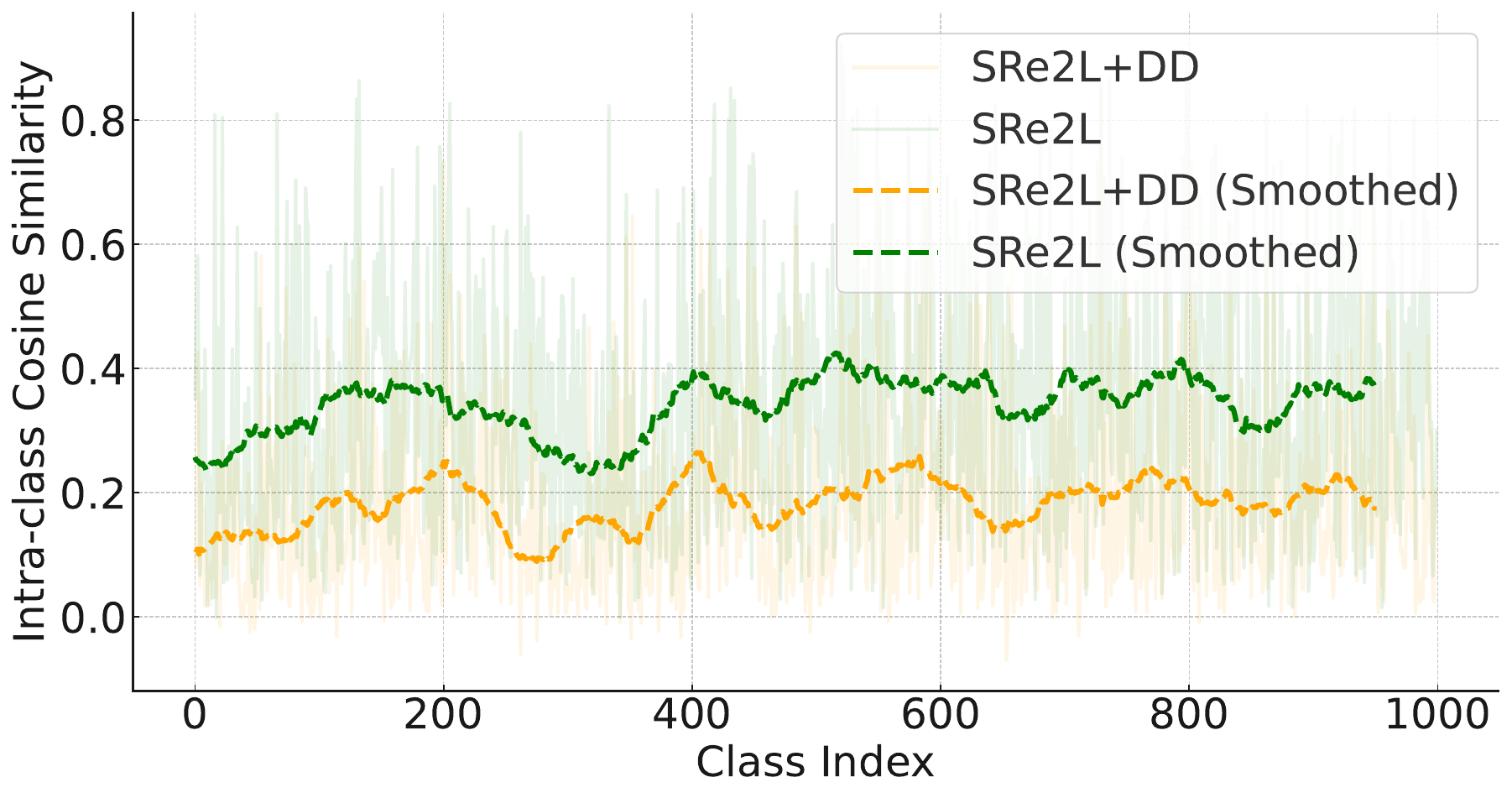}
\vspace{-15pt}
\end{center}
   \caption{\hll{Visualization of the mean cosine similarity between pairwise samples within the same class on ImageNet-1k under IPC 10.}}
\label{fig:motivation_of_dd}
\vspace{-12pt}
\end{figure}

\hll{As illustrated in Fig.~\ref{fig:motivation_of_dd}, the synthetic data generated by SRe2L exhibit excessive similarity within the same class, leading to a lack of diversity. Consequently, even if ``generalized matching'' preserves sufficient valuable information within a single image, the aggregate information content across all images within the same class does not increase effectively, which ultimately prevents ``generalized matching'' from being sufficiently advantageous. Data densification (DD) is proposed to address this by ensuring the data $\tilde{X}$ has full rank in the batch dimension, thereby guaranteeing that samples in each class are linearly independent, and ultimately ensuring that the data is diverse and abundant to fully exploit the potential of ``generalized matching''.}

To execute this pipeline, $\tilde{X}$ first needs to be downsampled to reduce the computational cost of eigenvalue decomposition:
\begin{equation}
\small
\begin{aligned}
& \hat{X} = \textrm{AvgPool2d}(\tilde{X}),\ s.t.\ \hat{X} \in \mathbb{R}^{B\times C \times 32 \times 32},\\
\end{aligned}
\label{eq:dd_1}
\end{equation}
where $B$ and $C$ represent the batch size and the number of channels, respectively. Afterward, we reshape $\hat{X}$ from ${B\!\times\! C\!\times\! 32 \!\times\! 32}$ to ${B \!\times\! (1024C)}$ and perform matrix multiplication in each class $y$ to obtain the set of the Gram matrix $\{\hat{X}_y\hat{X}_y^T\}_{y\in \mathcal{Y}}$, where $\mathcal{Y}$ refers to a set of all classes existing in this batch. And $\hat{X}_y$ is a subbatch with class $y$. Note that the alternative form $\{\hat{X}_y^T\hat{X}_y\}_{y\in \mathcal{Y}}$ is not desirable, as it is applicable only for dimensionality reduction in feature dimensions, which is why we do not consider singular value decomposition (SVD). To render $\hat{X}_y\hat{X}^T_y$ as full-rank as possible, we introduce the data densification loss in Eq.~\ref{eq:dd_2}.
\begin{equation}
\small
\begin{aligned}
& \mathcal{L}_\textrm{DD}(\tilde{X}) = \sum_{y\in \mathcal{Y}} D_\textrm{KL}(\textrm{stop\_grad}(p(\Sigma_y/\tau))||p(\Sigma_y)),\\
\end{aligned}
\label{eq:dd_2}
\end{equation}
where $\Sigma_y$, $\tau$, $\textrm{stop\_grad}(\cdot)$ and $p(\cdot)$ refer to the eigenvalues of $\hat{X}_y\hat{X}^T_y$, the temperature, the stop gradient operator and the softmax function, respectively. And $D_\textrm{KL}(\cdot||\cdot)$ denotes Kullback-Leibler divergence. As demonstrated in Sec.~\ref{sec:add_dd_explan}, the diversity of the data is significantly enhanced by the employment of Eq.~\ref{eq:dd_2}. In our experiment, $\tau$ is set as $4$ in default and we do not assign a deliberate weight (set to 1 by default) to $\mathcal{L}_\textrm{DD}$ because $\mathcal{L}_\textrm{DD} \equiv 0$ at the early 10\% of the iterations. In other words, DD is quite stable and the optimization objective $\operatorname*{arg\,min}_{\{\Sigma_y\}_{y\in \mathcal{Y}}} \mathcal{L}_\textrm{DD}$ is equivalent to $\operatorname*{arg\,min}_{\{\Sigma_y\}_{y\in \mathcal{Y}}}\sum_{y}\sigma^2(\Sigma_y)$.

\begin{figure}[!t]
\begin{center}
\includegraphics[width=0.85\linewidth,trim={0.0cm 0.0cm 0cm 0cm},clip]{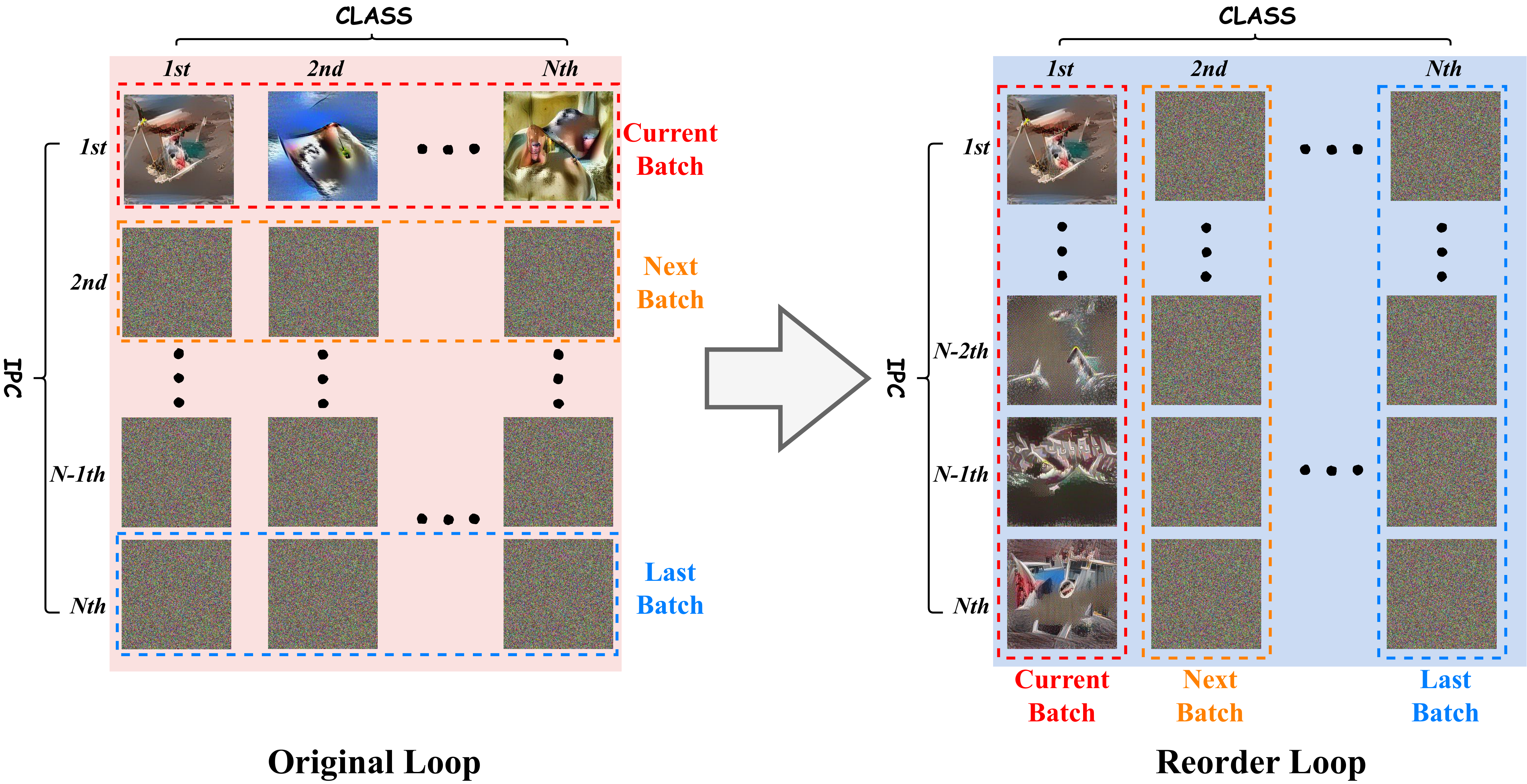}
\vspace{-20pt}
\end{center}
   \caption{The illustration of the original loop and the reorder loop.}
\label{fig:reorder_loop}
\vspace{-12pt}
\end{figure}

\paragraph{Technical Detail.} A problem that warrants attention is that in the SRe2L's implementation~\cite{dd_sre2l}, having merely a single sample in each class of a batch indicates insufficient to execute DD under the order of the original loop, as depicted in Fig.~\ref{fig:reorder_loop} (Left). A simple solution is to translate the original loop to the reorder loop, as shown in Fig.~\ref{fig:reorder_loop} (Right). However, our experiment on ResNet50 (\textit{i.e.}, the evaluation model) substantiates that this solution suffers a 2.6\% accuracy degradation (details can be found in Sec.~\ref{sec:ablation_studies}) on ImageNet-1k under IPC 10. The reason is that the number of classes in each iteration within the reorder loop is insufficient, preventing a batch
have the ability to match the global statistics in BatchNorm (\textit{i.e.}, ${\textbf{\textrm{BN}}_l^\textrm{CM}}$ and ${\textbf{\textrm{BN}}_l^\textrm{CV}}$). Motivated by score distillation sampling (SDS)~\cite{iclr2023_dreamfusion}, we update the statistics during data synthesis via EMA to solve this issue, so that the statistics of all past batches can assist the statistics of the current batch match ${\textbf{\textrm{BN}}_l^\textrm{CM}}$ and ${\textbf{\textrm{BN}}_l^\textrm{CV}}$:
\begin{equation}
\footnotesize
\begin{aligned}
& \mu_l^\textrm{total} = \alpha \mu_l^\textrm{total} + (1 - \alpha) \mu_l(\tilde{X}),\sigma^{2,\textrm{total}}_l= \alpha \sigma^{2,\textrm{total}}_l + (1 - \alpha) \sigma^2_l(\tilde{X}),\\
& \mathcal{L}_{\textrm{BN}}^{\prime}(\tilde{X}) = \sum_l || \mu_l(\tilde{X}) - {\textbf{\textrm{BN}}_l^\textrm{CM}} - \textrm{stop\_grad}(\mu_l(\tilde{X}) - \mu_l^\textrm{total})||_2 \\
& +|| \sigma^2_l(\tilde{X}) - {\textbf{\textrm{BN}}_l^\textrm{CV}} - \textrm{stop\_grad}(\sigma^2_l(\tilde{X}) - \sigma^{2,\textrm{total}}_l)||_2. \\
\end{aligned}
\label{eq:dd_3}
\end{equation}
The derivation of $\mathcal{L}_{\textrm{BN}}^{\prime}(\tilde{X})$ can be found in Appendix~\ref{apd:derivation_bn}. We call this lightweight strategy as ``match in the form of score distillation sampling'' and have demonstrated its effectiveness and feasibility in our ablation studies.

\subsection{Generalized Backbone Matching}
\label{sec:method_gbm}
\hll{Performing data synthesis only on a single pre-trained model is not able to enjoy ensemble gains from various backbones. Meanwhile, classical DC algorithms such as MTT~\cite{dd_mtt} and FTD~\cite{dd_minimizing_acc_traj_error} obtain performance improvements from  multiple randomly initialized backbones. Therefore, introducing generalized backbone matching (GBM) to apply various backbones for data synthesis is a desirable choice for ``generalized matching''.} It ensures a number of ``local-match-global'' matching nearly $N_\textrm{b}\times$ higher compared to just depending on a single backbone, where $N_\textrm{b}$ denotes the number of backbones. Regrettably, unrolling various backbone computational graphs in parallel for data synthesis is extremely expensive at the computational cost level. A solution is to randomly sample a backbone from the candidates in per iteration. This simple yet effective strategy not only ensures computational efficiency but also allows the data synthesis process to benefit from the diverse and multifaceted matching that the various backbones provide. We denote this pipeline as
\begin{equation}
\footnotesize
\begin{aligned}
& f_\textrm{cand}(\cdot) \sim \mathcal{U}(\mathbb{S}),\ \mathbb{S} = \{\textrm{ResNet18},\ \cdots,\ \textrm{ShuffleNetV2-0.5}\}.\\
\end{aligned}
\label{eq:gbm_1}
\end{equation}
To maintain backbone consistency in both the data synthesis and soft label generation phases, we introduce a backbone ensemble during soft label generation:
\begin{equation}
\tilde{z}\!=\!\left\{
\fontsize{8pt}{11pt}\selectfont\begin{aligned}
\sum_{f_\textrm{cand}\in\mathbb{S}}\frac{\sum_{g\in\mathbb{S}} ||g(\tilde{X})||_\textrm{F}}{|\mathbb{S}|||f_\textrm{cand}(\tilde{X})||_\textrm{F}}\frac{f_\textrm{cand}(\tilde{X})}{|\mathbb{S}|}&,\ & \textrm{w/ LN}, \\
\sum_{f_\textrm{cand}\in\mathbb{S}}\frac{f_\textrm{cand}(\tilde{X})}{|\mathbb{S}|}&,\ & \textrm{w/o LN}, \\
\end{aligned}
\right.
\label{eq:gbm_2}
\end{equation}
where $||\cdot||_\textrm{F}$ and LN refer to the Frobenius norm and Logit Normalization, respectively. Thus, we can obtain the soft label by $\tilde{y} = \textrm{softmax}(\tilde{z}/\tau)$. Particularly, the use of LN is optional, as demonstrated by our ablation studies; it is beneficial for ResNet18 but not for ResNet50 or ResNet101. Moreover, it's important to highlight that we apply a parallel mechanism for soft label generation since it provides a significantly lower computational cost, less than $1/30$ of that required for data synthesis, thus making the computational overhead negligible.

\subsection{Generalized Statistical Matching}
\label{sec:method_gsm}

\begin{figure}[!t]
\begin{center}
\includegraphics[width=0.95\linewidth,trim={0cm 0.2cm 0cm 0cm},clip]{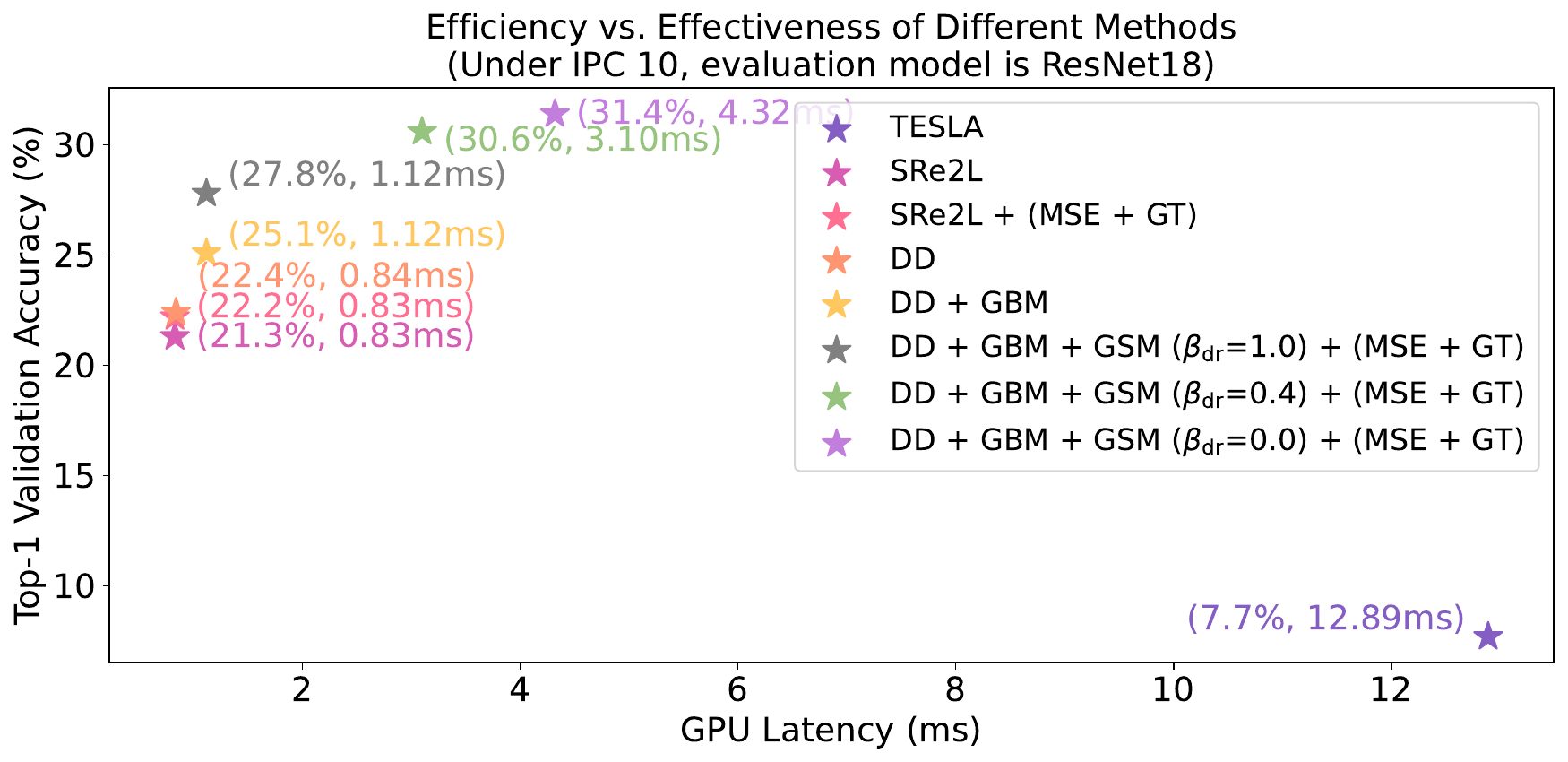}
\vspace{-10pt}
\end{center}
   \caption{Comparison of the effectiveness and efficiency of G-VBSM components. Among them, ``DD+GBM+GSM ($\beta_\textrm{dr}$=0.0)+(MSE+GT)'' represents the comprehensive G-VBSM.}
\label{fig:efficiency_vs_effectiveness}
\vspace{-12pt}
\end{figure}

Only ensuring backbone diversity is insufficient to fully exploit the potential of ``generalized matching''. In this subsection, we aim to introduce additional statistics for matching during the data synthesis phase. Prior methods~\cite{deepinversion,dd_sre2l} only utilize BatchNorm since the presence of global information statistics within the pre-trained model, with no apparent solution ready for other layers (\textit{e.g.}, Convolution). Retraining the model is the simplest way to address this problem, followed by updating the other layers' parameters through EMA. This approach is obviously impractical, as it defeats the purpose of using gradient descent. By contrast, we propose to allow the pre-trained model $f_\textrm{eval}(\cdot)$ to loop through the training dataset $\mathcal{T}$ once without calculating the gradient to obtain the global statistics of Convolution, thus serving ``local-match-global'':
\begin{equation}
\footnotesize
\begin{aligned}
&  \textbf{\textrm{Conv}}_l^\textbf{\textrm{CM}} = \frac{1}{|\mathcal{T}|}\sum_i^{|\mathcal{T}|} \textbf{\textrm{CM}}^l_i,\ \textbf{\textrm{Conv}}_l^\textbf{\textrm{CV}} = \frac{1}{|\mathcal{T}|} \sum_i^{|\mathcal{T}|}\textbf{\textrm{CV}}^l_i,\\
&  \textbf{\textrm{Conv}}_l^\textbf{\textrm{PM}} = \frac{1}{|\mathcal{T}|}\sum_i^{|\mathcal{T}|} \textbf{\textrm{PM}}^l_i,\ \textbf{\textrm{Conv}}_l^\textbf{\textrm{PV}} = \frac{1}{|\mathcal{T}|} \sum_i^{|\mathcal{T}|}\textbf{\textrm{PV}}^l_i.\\
\end{aligned}
\label{eq:gsm_1}
\end{equation}
Here, $\textbf{\textrm{CM}}^l_i\in \mathbb{R}^{C_l}$, $\textbf{\textrm{CV}}^l_i \in \mathbb{R}^{C_l}$, $\textbf{\textrm{PM}}^l_i \in \mathbb{R}^{\lceil \frac{H}{N^p_l} \rceil\!\times\!\lceil \frac{W}{N^p_l} \rceil }$ and $\textbf{\textrm{PV}}^l_i \in \mathbb{R}^{\lceil \frac{H}{N^p_l} \rceil\!\times\!\lceil \frac{W}{N^p_l} \rceil }$ refer to the channel mean, the channel variance, the patch mean and the patch variance, respectively, for the $l$-th Convolution when the input to $f_\textrm{cand}(\cdot)$ is the $i$-th batch, where $C_l$ and $\lceil \frac{H}{N^p_l} \rceil\!\times\!\lceil \frac{W}{N^p_l} \rceil $ denote the number of channels and patches of the $l$-th Convolution, respectively. We define $N^p_l$ as 4, 4, 4 and 16 by default on CIFAR-10, CIFAR-100, Tiny-ImageNet and ImageNet-1k, respectively. After obtaining the global channel mean $\textbf{\textrm{Conv}}_l^\textbf{\textrm{CM}}$, the global channel variance $\textbf{\textrm{Conv}}_l^\textbf{\textrm{CV}}$, the global patch mean $\textbf{\textrm{Conv}}_l^\textbf{\textrm{PM}}$ and the global patch variance $\textbf{\textrm{Conv}}_l^\textbf{\textrm{PV}}$, we can store them in a disk thus avoiding secondary calculations. In the data synthesis phase, we introduce $\mathcal{L}_\textrm{Conv}^{\prime}(\tilde{X})$ in Eq.~\ref{eq:gsm_2} to accomplish joint matching with $\mathcal{L}_\textrm{BN}^{\prime}(\tilde{X})$.
\begin{equation}
\fontsize{8pt}{10pt}\selectfont
\begin{aligned}
& \mathcal{L}_{\textrm{Conv}}^{\prime}(\tilde{X}) = \sum_l || \mu^c_l(\tilde{X}) - {\textbf{\textrm{Conv}}_l^\textrm{CM}} - \textrm{stop\_grad}(\mu^c_l(\tilde{X}) - \mu_l^{c,\textrm{total}})||_2 \\
&+ || \sigma^{c,2}_l(\tilde{X}) - {\textbf{\textrm{Conv}}_l^\textrm{CV}} - \textrm{stop\_grad}(\sigma^{c,2}_l(\tilde{X}) - \sigma^{c,2,\textrm{total}}_l)||_2, \\
 & + || \mu^p_l(\tilde{X}) - {\textbf{\textrm{Conv}}_l^\textrm{PM}} - \textrm{stop\_grad}(\mu^p_l(\tilde{X}) - \mu_l^{p,\textrm{total}})||_2 \\
 &+ || \sigma^{p,2}_l(\tilde{X}) - {\textbf{\textrm{Conv}}_l^\textrm{PV}} - \textrm{stop\_grad}(\sigma^{p,2}_l(\tilde{X}) - \sigma^{p,2,\textrm{total}}_l)||_2, \\
\end{aligned}
\label{eq:gsm_2}
\end{equation}
where $\mu_l^{c,\textrm{total}}$, $\sigma^{c,2,\textrm{total}}_l$, $\mu_l^{p,\textrm{total}}$, and $\sigma^{p,2,\textrm{total}}_l$ are each updated from the channel mean $\mu^c_l(\tilde{X})$, channel variance $\sigma^{c,2}_l(\tilde{X})$, patch mean $\mu^p_l(\tilde{X})$, and patch variance $\sigma^{p,2}_l(\tilde{X})$ respectively, all obtained via EMA from the current batch. In experiments, we discovered that Eq.~\ref{eq:gsm_2} causes a sightly larger computational burden, so we randomly drop the matching of statistics with a probability of $\beta_\textrm{dr}$ to ensure the efficiency of GSM.
\vspace{-10pt}
\paragraph{Loss Function in the Evaluation Phase.} \hll{Unlike DD, GBM and GSM are designed to create a distilled dataset that is enriched with information. Here, we introduce an enhancement to the loss function tailored specifically for the evaluation phase. Essentially, the evaluation phase is a knowledge distillation framework for transferring knowledge from a pre-trained model to the evaluation model.} SRe2L utilizes $D_\textrm{KL}(\tilde{y}||\textrm{softmax}(f_\textrm{eval}(\tilde{X})/\tau))$ as the loss function and experimentally illustrates that it improves performance by roughly 10\%. As established in SRe2L, an increase in temperature $\tau$ correlates with enhanced performance of the evaluation model. Inspired by this and $\tau^2 D_\textrm{KL}(\textrm{softmax}(p)/\tau||\textrm{softmax}(q)/\tau)$ is equivalent to $\frac{1}{2C}||p-q||_2^2$ when $\tau \rightarrow +\infty$~\cite{kim2021comparing}, we introduce a novel loss function MSE+$\gamma$$\times$GT to avoid numerical error caused by the large $\tau$ and improve the generalization of the distilled dataset without any additional overhead (ignore the weights $\tau^2$ and $\frac{1}{2C}$):
\begin{equation}
\footnotesize
\begin{aligned}
& \mathcal{L}_\textrm{eval}(\tilde{X},\tilde{y},y) = ||f_\textrm{eval}(\tilde{X})-\tilde{z}||_2^2 - \gamma \textbf{y}\log(\textrm{softmax}(f_\textrm{eval}(\tilde{X}))),\\
\end{aligned}
\label{eq:gbm_3}
\end{equation}
where $\textrm{y}$ represents the one-hot encoding (\textit{w.r.t.} the ground truth label $y$). As illustrated in Fig.~\ref{fig:efficiency_vs_effectiveness}, simply replacing the loss function with MSE+0.1$\times$GT ($\gamma$ is set as 0.1) in SRe2L improves the performance of ResNet18 (\textit{i.e.}, the evaluation model) by a margin of 0.9\% on ImageNet-1k under IPC 10.
\vspace{-10pt}
\section{Experiment}
\label{sec:experiment}

We perform comparison experiments on the large-scale dataset including the full 224$\times$224 ImageNet-1k~\cite{ILSVRC15} and the small-scale datasets including Tiny-ImageNet~\cite{tiny_imagenet}, CIFAR-10/100~\cite{CIFAR}. To highlight that our proposed G-VBSM \hll{is designed for} large-scale datasets, all ablation experiments are performed on ImageNet-1k.
\vspace{-5pt}
\paragraph{Hyperparameter Settings.} We prioritize selecting various convolution-based backbone, ensuring the maximal difference in architecture, while also adhering to the criterion of minimizing the number of parameters, an approach empirically demonstrated to be superior~\cite{dd_sre2l,dd_model_augmentation}. On ImageNet-1k, we skip the model pre-training phase by directly using Torchvision's open source pre-training weights~\cite{pytorch} of \{ResNet18~\cite{ResNet}, MobileNetV2~\cite{mobilenetv2}, EfficientNet-B0~\cite{EfficientNetV2}, ShuffleNetV2-0.5~\cite{shufflenet}\}. For the remaining dataset, we all train \{128-width ConvNet~\cite{shufflenet}, WRN-16-2~\cite{Zagoruyko2016WRN}, ResNet18~\cite{ResNet}, ShuffleNetV2-0.5~\cite{shufflenet}, MobileNetV2-0.5~\cite{mobilenetv2}\} from scratch with few epoch in the model pre-training phases. Gray \colorbox{gray!25}{cells} in all tables represent the highest performance. Meanwhile, (R18), (R50), etc. in all tables, represent the evaluation models. More details about the remaining method hyperparameter settings and additional ablation studies can be found in Appendix~\ref{apd:hyperparameter_setting} and Appendix~\ref{apd:imagenet_add} \&~\ref{apd:cifar_10_and_100_add}, respectively.

\subsection{Ablation Studies}
\label{sec:ablation_studies}

\paragraph{The Trade-off between Efficiency and Effectiveness.} The performance improvements and increased GPU latency for components of G-VBSM in comparison to the \baseline{baseline} SRe2L, are illustrated in Fig.~\ref{fig:efficiency_vs_effectiveness}. Clearly, DD, GBM, and GSM are highly effective in enhancing the generalization of distilled datasets. Compared with GSM, both DD and GBM are slightly lightweight and efficient. With DD and GBM alone \mbox{--} \textit{i.e.}, DD+GBM+GSM ($\beta_\textrm{dr}$=1.0)+(MSE+GT) \mbox{--} the accuracy of this approach surpasses that of SRe2L by a significant margin of 6.5\%. Additionally, the comprehensive G-VBSM, DD+GBM+GSM ($\beta_\textrm{dr}$=0.0)+(MSE+GT), further enhances the performance of DD+GBM+GSM ($\beta_\textrm{dr}$=1.0)+(MSE+GT) by a notable margin of 3.6\%. Note that MSE+$\gamma$$\times$GT ($\gamma$ is set as 0.1) is also extremely critical, for SRe2L and DD+GBM boosting 0.9\% and 2.7\%, respectively.
\begin{table}[h]
\small
\renewcommand\arraystretch{0.75}
\setlength\tabcolsep{0.85mm}
\begin{center}
\begin{tabular}{c|cc}
\shline
$\alpha$\verb|\|Evaluation Model & \makecell{ResNet18\\(MSE+0.1$\times$GT)} & \makecell{ResNet50\\(MSE+0.1$\times$GT)} \\
\hline
0.0 & 30.4\% & 31.9\% \\
0.4 & 31.4\% & 34.5\% \\
0.8 & \cellcolor{gray!25} 31.9\% & \cellcolor{gray!25} 36.4\% \\
\shline
\end{tabular}
\vspace{-5pt}
\end{center}
\caption{ImageNet-1k Top-1 Acc. about different $\alpha$ under IPC 10.}
\label{table:ab_alpha}
\vspace{-15pt}
\end{table}

\vspace{-10pt}
\paragraph{Matching in the Form of SDS.} Benefiting from this matching form, G-VBSM can achieve the best performance with small batch sizes (\textit{e.g.}, 40 in ImageNet-1k and 50 in CIFAR-100) in different settings, as demonstrated in our comparative experiments. Table~\ref{table:ab_alpha} demonstrates the impact of the factor $\alpha$ for performing the EMA update on the final performance achieved by G-VBSM, where $\alpha$=0 indicates that $\mathcal{L}^\prime_\textrm{BN}$ degenerates to $\mathcal{L}_\textrm{BN}$. We can conclude that the SDS matching form is feasible and the generalization of the distilled dataset improves with increasing $\alpha$.
\begin{table}[!h]
\small 
\renewcommand\arraystretch{0.95}
\setlength\tabcolsep{5pt}
\centering 
\resizebox{0.49\textwidth}{!}{%
\begin{tabular}{l|cccc}
\shline
\multirow{2}{*}{Method} & \multicolumn{4}{c}{Evaluation Model} \\
 & DeiT-Tiny & ResNet18 & MobileNetV2 & Swin-Tiny \\\hline
SRe2L & 15.41\% & 46.79\% & 36.59\% & 39.23\% \\
 G-VBSM & \cellcolor{gray!25} 29.43\% & \cellcolor{gray!25} 51.82\% & \cellcolor{gray!25} 48.66\% & \cellcolor{gray!25} 57.40\% \\
\shline
\end{tabular}}
\caption{ImageNet-1k Top-1 Acc. on cross-architecture generalization under IPC 50.}
\label{table:ab_cross_architecture}
\vspace{-15pt}
\end{table}

\vspace{-7pt}
\paragraph{Cross-Architecture Generalization.} The evaluation of cross-architecture generalization is crucial in assessing the quality of distilled datasets. Unlike traditional methods~\cite{dd_mtt,dd_kip,dd_datadam} which focus on CIFAR-100, our approach evaluates the effectiveness of the distilled dataset on ImageNet-1k, employing a suite of models with real-world applicability, including ResNet18~\cite{ResNet}, MobileNetV2~\cite{mobilenetv2}, DeiT-Tiny~\cite{deit}, and Swin-Tiny~\cite{SWIN-T}. The experimental results are reported in Table~\ref{table:ab_cross_architecture}. From Tables~\ref{tab:hyperparameter_imagenet_1k} and~\ref{table:ab_cross_architecture}, it is evident that the G-VBSM-synthetic dataset can effectively generalize across ResNet \{18, 50, 101\}, MobileNetV2, DeiT-Tiny, and Swin-Tiny architectures. Notably, DeiT-Tiny and Swin-Tiny, two architectures not encountered during the data synthesis phase, demonstrates significant proficiency with the accuracy 29.43\% and 57.40\%, outperforming the latest SOTA SRe2L by a margin of 14.01\% and 18.17\%, respectively.
\begin{table}[!h]
\footnotesize
\renewcommand\arraystretch{0.95}
\setlength\tabcolsep{0.7mm}
\begin{center}
\resizebox{0.49\textwidth}{!}{%
\begin{tabular}{cccc|cc}
\shline
\multicolumn{4}{c|}{Candidates (Backbone)} & \multicolumn{2}{c}{Evaluation Model} \\\hline
ResNet18 & MobileNetV2 & EfficientNet-B0 & ShuffleNetV2-0.5 & ResNet18 & ResNet50 \\
\checkmark & & & & 25.7\% & 30.2\% \\
\checkmark & \checkmark & & & 27.2\% & 31.2\% \\
\checkmark & \checkmark & \checkmark & & 27.9\% & 32.0\% \\
\checkmark & \checkmark & \checkmark & \checkmark &  \cellcolor{gray!25} 31.4\% &  \cellcolor{gray!25} 34.5\% \\
\shline
\end{tabular}}
\vspace{-10pt}
\end{center}
\caption{ImageNet-1k Top-1 Acc. about the number of candidate backbone in the soft label generation pipeline under IPC 10.}
\label{table:soft_label_various_backbone}
\vspace{-15pt}
\end{table}

\begin{figure*}[!t]
\begin{center}
\includegraphics[height=0.17\textheight,trim={0cm 0.4cm 0cm 0cm},clip]{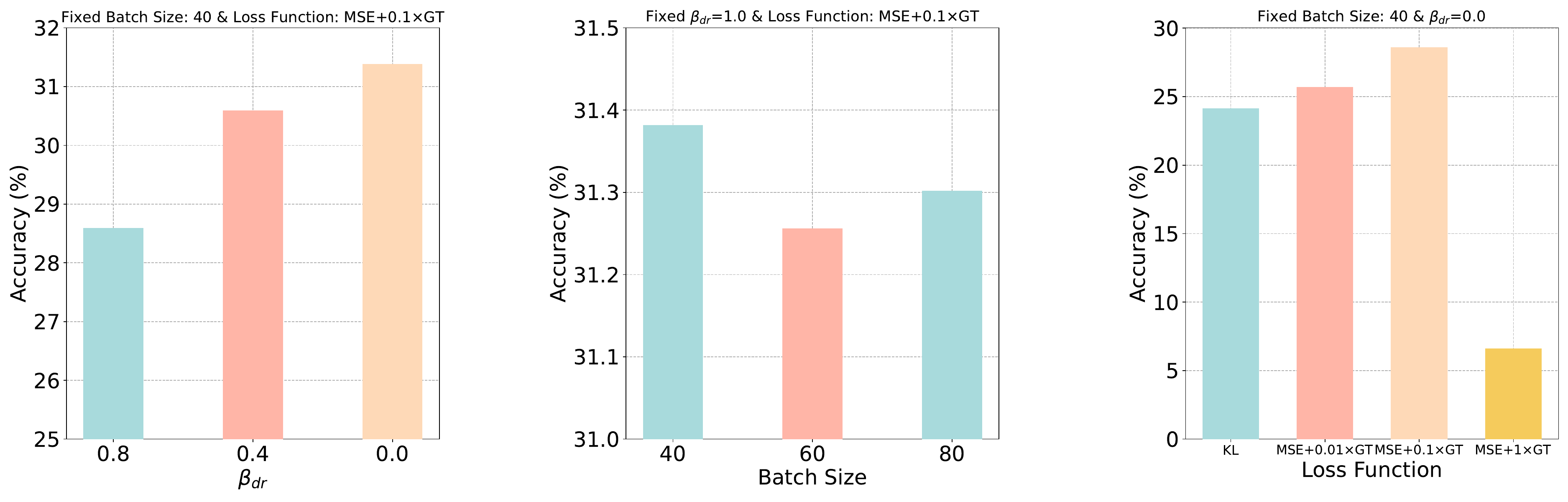}
\vspace{-15pt}
\end{center}
   \caption{ImageNet-1k Top-1 Acc. of different loss function, $\beta_\textrm{dr}$ and batch sizes under IPC 10.}
\label{fig:ablation_studies_loss_function}
\vspace{-5pt}
\end{figure*}
\begin{table*}[!t]
\centering
\resizebox{1.0\textwidth}{!}{
\begin{tabular}{ccccccccccc}
\toprule
Dataset                        & IPC & MTT~\cite{dd_mtt} (CW128) & DataDAM~\cite{dd_datadam} (CW128) & TESLA~\cite{dd_tesla} (R18) & SRe2L (R18) & SRe2L (R50) & SRe2L (R101) &  G-VBSM (R18) & G-VBSM (R50) & G-VBSM (R101) \\ \midrule
\multirow{2}{*}{Tiny-ImageNet}                 & 50  &  28.0$\pm$0.3  & 28.7$\pm$0.3 &   -  & 41.1$\pm$0.4 & 42.2$\pm$0.5 &  42.5$\pm$0.2  &  \cellcolor{gray!25} 47.6$\pm$0.3 & \cellcolor{gray!25} 48.7$\pm$0.2 &  \cellcolor{gray!25} 48.8$\pm$0.4  \\ 
 & 100  & -   &  -    & -  & 49.7$\pm$0.3 &51.2$\pm$0.4 & 51.5$\pm$0.3 & \cellcolor{gray!25} 51.0$\pm$0.4 & \cellcolor{gray!25} 52.1$\pm$0.3 & \cellcolor{gray!25} 52.3$\pm$0.1 \\\midrule
\multirow{3}{*}{\makecell{ImageNet-1k}}  & 10  & \cellcolor{gray!25} 64.0$\pm$1.3$^\dag$ & 6.3$\pm$0.0 & 7.7$\pm$0.1    &  21.3$\pm$0.6 &  28.4$\pm$0.1 &  30.9$\pm$0.1 &  \cellcolor{gray!25} 31.4$\pm$0.5 &  \cellcolor{gray!25} 35.4$\pm$0.8 &  \cellcolor{gray!25} 38.2$\pm$0.4  \\
  & 50  &  -  & 15.5$\pm$0.2 &  -  & 46.8$\pm$0.2    & 55.6$\pm$0.3    & 60.8$\pm$0.5 & \cellcolor{gray!25} 51.8$\pm$0.4  &  \cellcolor{gray!25} 58.7$\pm$0.3 &  \cellcolor{gray!25} 61.0$\pm$0.4 \\ 
  & 100 &  -  & - & -  & 52.8$\pm$0.3 & 61.0$\pm$0.4 & 62.8$\pm$0.2 & \cellcolor{gray!25} 55.7$\pm$0.4 & \cellcolor{gray!25} 62.2$\pm$0.3 & \cellcolor{gray!25} 63.7$\pm$0.2 \\ 
    \bottomrule
\end{tabular}
}
\caption{Comparison with baseline models in Tiny-ImageNet and ImageNet-1k. $^\dag$ indicates the ImageNette dataset, which contains only 10 classes. DataDAM~\cite{dd_datadam} and TESLA~\cite{dd_tesla} use the downsampled 64$\times$64 ImageNet-1k. We cite the experimental results from SRe2L~\cite{dd_sre2l}.}
\label{tab:baseline_final}
\vspace{-0.1in}
\end{table*}

\begin{table*}[!t]
\renewcommand\arraystretch{1}
\centering
\scriptsize
\setlength{\tabcolsep}{1.65pt}
\setlength{\abovecaptionskip}{0.1cm}
\resizebox{1\linewidth}{!}{
\begin{tabular}{cc|cccc|ccccccccc|c}
\toprule
\multirow{2}{*}{Dataset}           & \multirow{2}{*}{IPC} & \multicolumn{4}{c|}{Coreset Selection}   & \multicolumn{7}{c|}{Training Set Synthesis (CW128)} & \multicolumn{2}{c|}{Training Set Synthesis (R18)} & \multirow{2}{*}{\makecell{Whole Dataset\\(CW128)}} \\ %
                            & &              
                            \multicolumn{1}{c}{Random}        & \multicolumn{1}{c}{Herding}       & \multicolumn{1}{c}{K-Center}      & \multicolumn{1}{c|}{Forgetting}   &          \multicolumn{1}{c}{{DC} \cite{dd_gradient_matching}}          	&	
                            \multicolumn{1}{c}{{DM} \cite{dd_dist_matching}}         &  \multicolumn{1}{c}{{CAFE} \cite{dd_CAFE}}         &\multicolumn{1}{c}{{KIP} \cite{dd_kip}} &  
                            \multicolumn{1}{c}{{MTT} \cite{dd_mtt}} &\multicolumn{1}{c}{DataDAM\cite{dd_datadam}} & \multicolumn{1}{c|}{G-VBSM} & SRe2L  &  \multicolumn{1}{c|}{G-VBSM} 
                              \\ \midrule

\multirow{2}{*}{CIFAR-10}  &  10  & 26.0$\pm$1.2  & 31.6$\pm$0.7 & 14.7$\pm$0.9 & 23.3$\pm$1.0 & 44.9$\pm$0.5 & 48.9$\pm$0.6 & 50.9$\pm$0.5 & 46.1$\pm$0.7 & \cellcolor{gray!25} 65.3$\pm$0.7 & \multicolumn{1}{c}{54.2$\pm$0.8} & \multicolumn{1}{c|}{46.5$\pm$0.7} &  \multicolumn{1}{c}{27.2$\pm$0.5} & \multicolumn{1}{c|}{\cellcolor{gray!25} 53.5$\pm$0.6} & \multirow{2}{*}{84.8$\pm$0.1} \\

& 50  & 43.4$\pm$1.0  & 40.4$\pm$0.6 & 27.0$\pm$1.4 & 23.3$\pm$1.1  & 53.9$\pm$0.5 & 63.0$\pm$0.4 & 62.3$\pm$0.4 & 53.2$\pm$0.7 & \cellcolor{gray!25} 71.6$\pm$0.2 &\multicolumn{1}{c}{\cellcolor{gray!25} 67.0$\pm$0.4} & \multicolumn{1}{c|}{54.3$\pm$0.3} & \multicolumn{1}{c}{47.5$\pm$0.6} & \multicolumn{1}{c|}{\cellcolor{gray!25} 59.2$\pm$0.4}  & \\ 
\midrule
                                                            
\multirow{3}{*}{CIFAR-100} & 1  &  4.2$\pm$0.3  & 8.3$\pm$0.3 & 8.4$\pm$0.3  &  4.5$\pm$0.2 & 12.8$\pm$0.3 & 11.4$\pm$0.3 & 14.0$\pm$0.3 & 12.0$\pm$0.2 & \cellcolor{gray!25} 24.3$\pm$0.3 & \multicolumn{1}{c}{14.5$\pm$0.5} & \multicolumn{1}{c|}{ 16.4$\pm$0.7} &  \multicolumn{1}{c}{2.0$\pm$0.2} & \multicolumn{1}{c|}{\cellcolor{gray!25} 25.9$\pm$0.5} & \multirow{3}{*}{56.2$\pm$0.3} \\
                              
& 10 & 14.6$\pm$0.5  & 17.3$\pm$0.3  & 17.3$\pm$0.3 & 15.1$\pm$0.3 & 25.2$\pm$0.3 & 29.7$\pm$0.3  & 31.5$\pm$0.2 & 40.1$\pm$0.4 & 33.1$\pm$0.4 & \multicolumn{1}{c}{34.8$\pm$0.5} & \multicolumn{1}{c|}{ \cellcolor{gray!25} 38.7$\pm$0.2} &  \multicolumn{1}{c}{ 31.6$\pm$0.5}  & \multicolumn{1}{c|}{\cellcolor{gray!25} 59.5$\pm$0.4} & \\

& 50 & 30.0$\pm$0.4  & 33.7$\pm$0.5  & 30.5$\pm$0.3  & \multicolumn{1}{c|}{-} & 30.6$\pm$0.6  &  43.6$\pm$0.4  & 47.7$\pm$0.2 & -  & 42.9$\pm$0.3 & \multicolumn{1}{c}{\cellcolor{gray!25} 49.4$\pm$0.3} & \multicolumn{1}{c|}{45.7$\pm$0.4} & \multicolumn{1}{c}{49.5$\pm$0.3}  & \multicolumn{1}{c|}{\cellcolor{gray!25} 65.0$\pm$0.5} & \\\bottomrule
\end{tabular}}
\caption{Comparison with baseline models on CIFAR-10/100. All methods, except for SRe2L and G-VBSM, use a 128-width ConvNet (CW128) for data synthesis and evaluation. G-VBSM utilizes \{CW128, WRN-16-2, ResNet18 (R18), ShuffleNetV2-0.5, MobileNetV2-0.5\} for data synthesis and \{CW128, R18\} for evaluation. We cite the experimental results, except for SRe2L's, from DataDAM~\cite{dd_datadam}.} 
\label{tab:baseline_traditional}
\vspace{-10pt}
\end{table*}
\vspace{-10pt}
\paragraph{Ensemble in Soft Label Generation.} In knowledge distillation, soft labels obtained through multiple teacher ensembles can effectively enhance the generalization of the student~\cite{multi_teacher}. This observation is similarly corroborated in dataset distillation, as evidenced in Table~\ref{table:soft_label_various_backbone}. The greater the number of models, the stronger the generalization of the distilled dataset. It is particularly interesting to point out that ShuffleNetV2-0.5 improves further by more than 2\% even though the other three models have already achieved modest ensemble gains. We attribute this enhancement to the channel shuffle mechanism within ShuffleNetV2-0.5, which imposes a beneficial regularization constraint for G-VBSM.
\begin{table}[!h]
\vspace{-10pt}
\centering
\footnotesize
\renewcommand\arraystretch{0.95}
\resizebox{0.49\textwidth}{!}{%
\begin{tabular}{c|cccc}
\shline
\multirow{3}{*}{\makecell{Logit\\Normalization}} & \multicolumn{4}{c}{Evaluation Model} \\
& \makecell{ResNet18\\(MSE+0.1$\times$GT)} & \makecell{ResNet18\\(KL)} & \makecell{ResNet50\\(MSE+0.1$\times$GT)} & \makecell{ResNet50\\(KL)} \\\hline
Yes & \cellcolor{gray!25} 31.4\% & \cellcolor{gray!25} 25.4\% & 34.5\% & 28.6\% \\
No & 31.0\% & 25.1\% & \cellcolor{gray!25} 35.4\% & \cellcolor{gray!25} 31.9\% \\
\shline
\end{tabular}}
\caption{ImageNet-1k Top-1 Acc. about the use of logit normalization under IPC 10.}
\label{table:bn_logit_normalization}
\vspace{-10pt}
\end{table}

\vspace{-15pt}
\paragraph{Logit Normalization in Soft Label Generation.} The aim of this strategy is to maintain the consistency of the logits (can be viewed as vectors) magnitude across all models within a high-dimensional space, thereby ensuring equal contribution of these models to the ultimate soft label for ensemble. This approach is not universally effective, as shown in Table~\ref{table:bn_logit_normalization}. When the evaluation model parameter count is low (\textit{e.g.}, ResNet18), it can enhance G-VBSM performance. Conversely, with a model like ResNet50, it may hinder performance, rendering G-VBSM uncompetitive. As a result, in this work, ResNet \{50, 101\} and ViT-based models do not employ logit normalization, while the remaining models do.
\vspace{-10pt}
\paragraph{Loss Function in the Evaluation Phase.} As illustrated in Fig.~\ref{fig:ablation_studies_loss_function}, replacing $D_\textrm{KL}(\cdot||\cdot)$ in SRe2L with MSE+$\gamma$$\times$GT demonstrates to be extremely effective. For example, with $\beta_\textrm{dr}$=0.0 and batch size=40, adjusting $\gamma$ to 0.01 and 0.1 enhances model performance during the evaluation phase by 1.4\% and 6.0\%, respectively. As a result, we employ a $\gamma$ of 0.1 in all experiments conducted on ImageNet-1k. Furthermore, we observe that the distilled dataset's generalization is not significantly affected ($\leq$2\%) by variations in either batch size or $\beta_\textrm{dr}$. This ensured that G-VBSM attain SOTA results with a minimal batch size 40 and $\beta_\textrm{dr}$=0.4 (\textit{i.e.}, the default settings in ImageNet-1k).

\begin{figure}[!h]
\begin{center}
\includegraphics[width=1.0\linewidth,trim={0cm 0.0cm 0cm 0cm},clip]{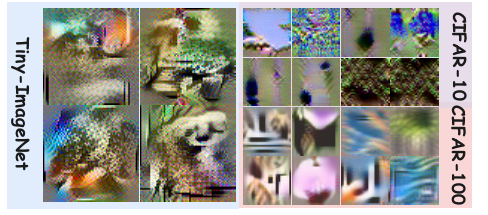}
\vspace{-25pt}
\end{center}
   \caption{Synthetic data visualization on small-scale datasets.}
\label{fig:visualization_toy_dataset}
\vspace{-15pt}
\end{figure}
\vspace{-5pt}
\paragraph{Synthetic Data Visualization.} The visualization results for the large-scale (the full 224$\times$224 ImageNet-1k) and small-scale (CIFAR-10/100 and Tiny-ImageNet) datasets are displayed in Figs.~\ref{fig:first_presentation} and~\ref{fig:visualization_toy_dataset}, respectively. On ImageNet-1k, the images synthesized by G-VBSM are more informative and abstract compared with those from SRe2L. Furthermore, on small-scale datasets, the distilled images obtained by G-VBSM markedly differ from the images presented in MTT and DataDAM's papers~\cite{dd_mtt,dd_datadam}, illustrating that G-VBSM is an out-of-the ordinary algorithm. More synthetic data can be found in Appendix~\ref{apd:visualization}.
\begin{figure}[!ht]
\centering
 \includegraphics[width=0.48\textwidth,trim={0cm 0cm 0cm 0cm},clip]{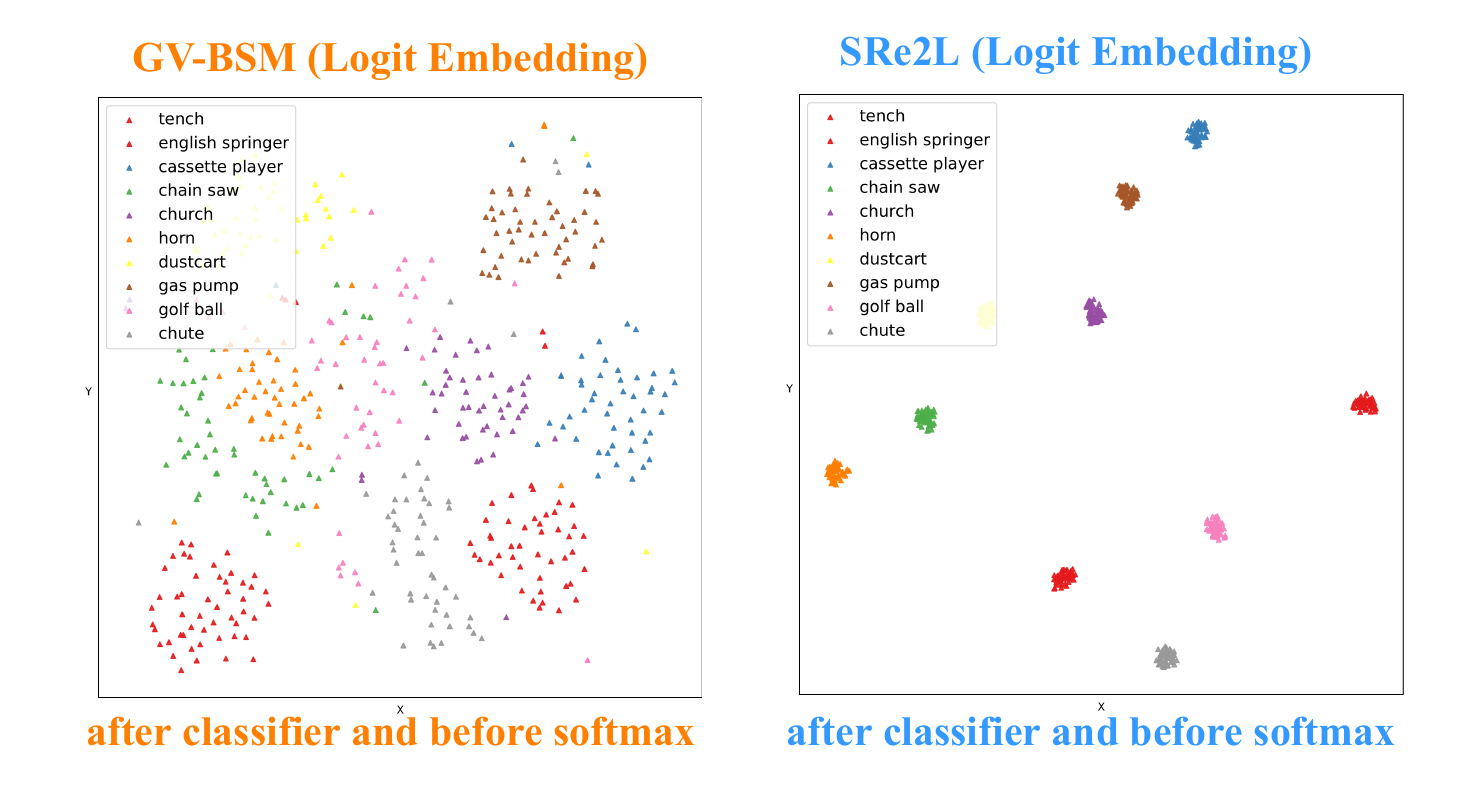}
  \vspace{-15pt}
 \caption{t-SNE visualization on ImageNet-1k.}
 \label{fig:t_sne}
 \vspace{-15pt}
\end{figure}
\paragraph{Logit Embedding Distributions Visualization.} We feed distilled datasets into a pre-trained model (\textit{i.e.}, ResNet18) to obtain logit embeddings for t-SNE~\cite{van2008visualizing} visualization. Note that unlike previous methods that used feature embeddings before the classifier layer, we utilize logit embeddings after the classifier layer for presentation. Fig.~\ref{fig:t_sne} shows the distributions of G-VBSM and SRe2L logit embeddings. The logit embedding obtained by G-VBSM has less intra-class similarity than SRe2L, ensuring softer labels, preventing model overconfidence during the evaluation phase, and ultimately enhancing model generalization.

\subsection{Large-Scale Dataset Comparison} 
\paragraph{Full 224$\times$224 ImageNet-1k.} As illustrated in the ImageNet-1k part of Table~\ref{tab:baseline_final}, G-VBSM consistently outperforms SRe2L across all IPC \{10, 50, 100\}, with the evaluation model being ResNet \{18, 50, 101\}, indicating that G-VBSM is highly competitive in the large-scale dataset ImageNet-1k. In particular, under extreme compression scenarios, such as IPC 10, G-VBSM achieves significant performance increases \mbox{--} 10.1\%, 7.0\%, and 7.3\% for ResNet18, ResNet50, and ResNet101, respectively \mbox{--} compared with SRe2L. In comparison to the latest SOTA classical method, DataDAM, just experimenting on the full 64$\times$64 ImageNet-1k, we exceed it by a margin of 23.7\% under IPC 10. These results are particularly impressive when considered in the context of the large-scale dataset ImageNet-1k.

\subsection{Small-Scale Dataset Comparison}
\paragraph{Tiny-ImageNet.} Similar to SRe2L, we evaluate the generalization of the distilled dataset on ResNet \{18, 50, 101\}. As presented in the Tiny-ImageNet part of Table~\ref{tab:baseline_final}, G-VBSM consistently surpasses the latest SOTA method SRe2L on different settings. Our method outperforms SRe2L, MTT and DataDAM by 6.5\%, 19.6\% and 18.9\%, respectively, under IPC 50 applying 128-width ConvNet or ResNet18 as the evaluation model.
\vspace{-6pt}
\paragraph{CIFAR-100.} One of the disadvantages of SRe2L is its inability to compete with SOTA performance on this small-scale dataset benchmark (\textit{i.e.}, evaluated with 128-width ConvNet and ResNet18), such as the performance on CIFAR-100 shown in Table~\ref{tab:baseline_traditional}, as a result, it fails to demonstrate that ``local-match-global'' (\textit{e.g.}, statistical matching of BatchNorm) can rival or outperform traditional DC algorithms. In contrast, as shown in the CIFAR-100 part of Table~\ref{tab:baseline_traditional}, G-VBSM achieves accuracies 16.4\% and 38.7\%, the highest under IPC 1 and 10 on CIFAR-100 through ``generalized matching'', while maintaining the same number of epochs (\textit{i.e.}, 1000) and the model (\textit{i.e.}, 128-width ConvNet) for the evaluation phase as the traditional methods, outperforming the latest SOTA DataDAM by 1.9\% and 3.9\%, respectively. To the best of our knowledge, our proposed G-VBSM is the first algorithm that is strongly competitive on CIAR-100 as well as the full 224$\times$224 ImageNet-1k.
\vspace{-6pt}
\paragraph{CIFAR-10.} As illustrated in the CIFAR-10 part of Table~\ref{tab:baseline_traditional}, G-VBSM outperforms vanilla DC~\cite{dd_begin}, KIP~\cite{dd_kip}, and all coreset selection algorithms on the benchmark with 128-width ConvNet as the evaluation model. Meanwhile, on the benchmark with ResNet 18 as the evaluation model, G-VBSM outperforms the \baseline{baseline} SRe2L by 26.3\% and 11.7\% under IPC 10 and 50, respectively.

\section{Conclusion}
\label{sec:conclusion}
\hll{In this paper, we introduce a novel perspective termed ``generalized matching'' for dataset condensation. This posits that an abundance of lightweight ``local-match-global'' matching are more effective than a single ``local-match-global'' matching, and even surpass the precise and costly matching used by traditional methods.} Consequently, we present G-VBSM, designed to ensure data densification while performing matching based on sufficient and various backbones, layers, and statistics. Experiments conducted on CIFAR-10/100, Tiny-ImageNet, and the full 224$\times$224 ImageNet-1k demonstrated that our method is the first algorithm to show strong performance across both small-scale and large-scale datasets. For future research, we plan to extend this approach's applicability to large-scale detection and segmentation datasets.
\clearpage

{
    \small
    \bibliographystyle{ieeenat_fullname}
    \bibliography{main}

\begin{thebibliography}{44}
\providecommand{\natexlab}[1]{#1}
\providecommand{\url}[1]{\texttt{#1}}
\expandafter\ifx\csname urlstyle\endcsname\relax
  \providecommand{\doi}[1]{doi: #1}\else
  \providecommand{\doi}{doi: \begingroup \urlstyle{rm}\Url}\fi

\bibitem[Cazenavette et~al.(2022)Cazenavette, Wang, Torralba, Efros, and Zhu]{dd_mtt}
George Cazenavette, Tongzhou Wang, Antonio Torralba, Alexei~A. Efros, and Jun{-}Yan Zhu.
\newblock Dataset distillation by matching training trajectories.
\newblock In \emph{Computer Vision and Pattern Recognition}, New Orleans, LA, USA, 2022. {IEEE}.

\bibitem[Cui et~al.(2023)Cui, Wang, Si, and Hsieh]{dd_tesla}
Justin Cui, Ruochen Wang, Si Si, and Cho{-}Jui Hsieh.
\newblock Scaling up dataset distillation to imagenet-1k with constant memory.
\newblock In \emph{International Conference on Machine Learning}, pages 6565--6590, Honolulu, Hawaii, {USA}, 2023. {PMLR}.

\bibitem[Du et~al.(2023)Du, Jiang, Tan, Zhou, and Li]{dd_minimizing_acc_traj_error}
Jiawei Du, Yidi Jiang, Vincent Y.~F. Tan, Joey~Tianyi Zhou, and Haizhou Li.
\newblock Minimizing the accumulated trajectory error to improve dataset distillation.
\newblock In \emph{Computer Vision and Pattern Recognition}, pages 3749--3758, Vancouver, BC, Canada, 2023. {IEEE}.

\bibitem[Gou et~al.(2021)Gou, Yu, Maybank, and Tao]{kdsurvey}
Jianping Gou, Baosheng Yu, Stephen~J Maybank, and Dacheng Tao.
\newblock Knowledge distillation: A survey.
\newblock \emph{International Journal of Computer Vision}, 129\penalty0 (6):\penalty0 1789--1819, 2021.

\bibitem[He et~al.(2016)He, Zhang, and Ren]{ResNet}
Kaiming He, Xiangyu Zhang, and Shaoqing Ren.
\newblock Deep residual learning for image recognition.
\newblock In \emph{Computer Vision and Pattern Recognition}, pages 770--778, Las Vegas, NV, USA, 2016. IEEE.

\bibitem[Hinton et~al.(2015)Hinton, Vinyals, and Dean]{vanillakd}
Geoffrey Hinton, Oriol Vinyals, and Jeff Dean.
\newblock Distilling the knowledge in a neural network, 2015.

\bibitem[Hu et~al.(2022)Hu, Goetz, Malik, Zhan, Liu, and Liu]{dd_federated_learning_1}
Shengyuan Hu, Jack Goetz, Kshitiz Malik, Hongyuan Zhan, Zhe Liu, and Yue Liu.
\newblock Fedsynth: Gradient compression via synthetic data in federated learning.
\newblock \emph{arXiv preprint arXiv:2204.01273}, 2022.

\bibitem[Huang et~al.(2022)Huang, You, Wang, Qian, and Xu]{DIST}
Tao Huang, Shan You, Fei Wang, Chen Qian, and Chang Xu.
\newblock Knowledge distillation from a stronger teacher.
\newblock In \emph{Neural Information Processing Systems}, 2022.

\bibitem[Kim et~al.(2022)Kim, Kim, Oh, Yun, Song, Jeong, Ha, and Song]{dd_efficient_parameterization}
Jang{-}Hyun Kim, Jinuk Kim, Seong~Joon Oh, Sangdoo Yun, Hwanjun Song, Joonhyun Jeong, Jung{-}Woo Ha, and Hyun~Oh Song.
\newblock Dataset condensation via efficient synthetic-data parameterization.
\newblock In \emph{International Conference on Machine Learning}, pages 11102--11118, Baltimore, Maryland, {USA}, 2022. {PMLR}.

\bibitem[Kim et~al.(2021)Kim, Oh, Kim, Cho, and Yun]{kim2021comparing}
Taehyeon Kim, Jaehoon Oh, NakYil Kim, Sangwook Cho, and Se-Young Yun.
\newblock Comparing kullback-leibler divergence and mean squared error loss in knowledge distillation.
\newblock In \emph{International Joint Conference on Artificial Intelligence}, Virtual Event, 2021. Morgan Kaufmann.

\bibitem[Krizhevsky et~al.(2009)Krizhevsky, Hinton, et~al.]{CIFAR}
Alex Krizhevsky, Geoffrey Hinton, et~al.
\newblock Learning multiple layers of features from tiny images.
\newblock 2009.

\bibitem[Liu et~al.(2023)Liu, Gu, Wang, Zhu, Jiang, and You]{dd_dream}
Yanqing Liu, Jianyang Gu, Kai Wang, Zheng Zhu, Wei Jiang, and Yang You.
\newblock {DREAM:} efficient dataset distillation by representative matching.
\newblock \emph{arXiv preprint arXiv:2302.14416}, 2023.

\bibitem[Liu et~al.(2021)Liu, Lin, Cao, Hu, Wei, Zhang, Lin, and Guo]{SWIN-T}
Ze Liu, Yutong Lin, Yue Cao, Han Hu, Yixuan Wei, Zheng Zhang, Stephen Lin, and Baining Guo.
\newblock Swin transformer: Hierarchical vision transformer using shifted windows.
\newblock In \emph{International Conference on Computer Vision}, pages 10012--10022, 2021.

\bibitem[Masarczyk and Tautkute(2020)]{dd_continual_learning_1}
Wojciech Masarczyk and Ivona Tautkute.
\newblock Reducing catastrophic forgetting with learning on synthetic data.
\newblock In \emph{Computer Vision and Pattern Recognition Workshops}, pages 252--253, Virtual Event, 2020. {IEEE}.

\bibitem[Nguyen et~al.(2020)Nguyen, Chen, and Lee]{dd_kip}
Timothy Nguyen, Zhourong Chen, and Jaehoon Lee.
\newblock Dataset meta-learning from kernel ridge-regression.
\newblock \emph{arXiv preprint arXiv:2011.00050}, 2020.

\bibitem[Paszke et~al.(2019)Paszke, Gross, Massa, Lerer, Bradbury, Chanan, Killeen, Lin, Gimelshein, Antiga, et~al.]{pytorch}
Adam Paszke, Sam Gross, Francisco Massa, Adam Lerer, James Bradbury, Gregory Chanan, Trevor Killeen, Zeming Lin, Natalia Gimelshein, Luca Antiga, et~al.
\newblock Pytorch: An imperative style, high-performance deep learning library.
\newblock In \emph{Neural Information Processing Systems}, Vancouver, BC, Canada, 2019.

\bibitem[Poole et~al.(2023)Poole, Jain, Barron, and Mildenhall]{iclr2023_dreamfusion}
Ben Poole, Ajay Jain, Jonathan~T. Barron, and Ben Mildenhall.
\newblock Dreamfusion: Text-to-3d using 2d diffusion.
\newblock In \emph{International Conference on Learning Representations}, 2023.

\bibitem[Russakovsky et~al.(2015)Russakovsky, Deng, Su, Krause, Satheesh, Ma, Huang, Karpathy, Khosla, Bernstein, et~al.]{ILSVRC15}
Olga Russakovsky, Jia Deng, Hao Su, Jonathan Krause, Sanjeev Satheesh, Sean Ma, Zhiheng Huang, Andrej Karpathy, Aditya Khosla, Michael Bernstein, et~al.
\newblock Imagenet large scale visual recognition challenge.
\newblock \emph{International Journal of Computer Vision}, 115\penalty0 (3):\penalty0 211--252, 2015.

\bibitem[Sachdeva and McAuley(2023)]{dd_survey}
Noveen Sachdeva and Julian McAuley.
\newblock Data distillation: A survey.
\newblock \emph{arXiv preprint arXiv:2301.04272}, 2023.

\bibitem[Sajedi et~al.(2023)Sajedi, Khaki, Amjadian, Liu, Lawryshyn, and Plataniotis]{dd_datadam}
Ahmad Sajedi, Samir Khaki, Ehsan Amjadian, Lucy~Z Liu, Yuri~A Lawryshyn, and Konstantinos~N Plataniotis.
\newblock Datadam: Efficient dataset distillation with attention matching.
\newblock In \emph{International Conference on Computer Vision}, pages 17097--17107, Paris, France, 2023. {IEEE}.

\bibitem[Sandler et~al.(2018)Sandler, Howard, Zhu, Zhmoginov, and Chen]{mobilenetv2}
Mark Sandler, Andrew~G. Howard, Menglong Zhu, Andrey Zhmoginov, and Liang{-}Chieh Chen.
\newblock Mobilenetv2: Inverted residuals and linear bottlenecks.
\newblock In \emph{Computer Vision and Pattern Recognition}, pages 4510--4520, Salt Lake City, UT, USA, 2018. IEEE.

\bibitem[Sangermano et~al.(2022)Sangermano, Carta, Cossu, and Bacciu]{dd_continual_learning_2}
Mattia Sangermano, Antonio Carta, Andrea Cossu, and Davide Bacciu.
\newblock Sample condensation in online continual learning.
\newblock In \emph{International Joint Conference on Neural Networks}, pages 1--8, Padua, Italy, 2022. {IEEE}.

\bibitem[Shen and Xing(2022)]{shen2022fast}
Zhiqiang Shen and Eric Xing.
\newblock A fast knowledge distillation framework for visual recognition.
\newblock In \emph{European Conference on Computer Vision}, pages 673--690. Springer, 2022.

\bibitem[Such et~al.(2020)Such, Rawal, Lehman, Stanley, and Clune]{dd_nas_1}
Felipe~Petroski Such, Aditya Rawal, Joel Lehman, Kenneth~O. Stanley, and Jeffrey Clune.
\newblock Generative teaching networks: Accelerating neural architecture search by learning to generate synthetic training data.
\newblock In \emph{International Conference on Machine Learning}, pages 9206--9216, Virtual Event, 2020. {PMLR}.

\bibitem[Tan and Le(2021)]{EfficientNetV2}
Mingxing Tan and Quoc Le.
\newblock Efficientnetv2: Smaller models and faster training.
\newblock In \emph{International Conference on Machine Learning}, pages 10096--10106, Virtual Event, 2021. PMLR.

\bibitem[Tavanaei(2020)]{tiny_imagenet}
Amirhossein Tavanaei.
\newblock Embedded encoder-decoder in convolutional networks towards explainable {AI}.
\newblock 2020.

\bibitem[Tian et~al.(2019)Tian, Krishnan, and Isola]{CRD}
Yonglong Tian, Dilip Krishnan, and Phillip Isola.
\newblock Contrastive representation distillation.
\newblock In \emph{International Conference on Learning Representations}, 2019.

\bibitem[Touvron et~al.(2021)Touvron, Cord, Douze, Massa, Sablayrolles, and J{\'{e}}gou]{deit}
Hugo Touvron, Matthieu Cord, Matthijs Douze, Francisco Massa, Alexandre Sablayrolles, and Herv{\'{e}} J{\'{e}}gou.
\newblock Training data-efficient image transformers {\&} distillation through attention.
\newblock In \emph{International Conference on Machine Learning}, pages 10347--10357, Virtual Event, 2021. {PMLR}.

\bibitem[Van~der Maaten and Hinton(2008)]{van2008visualizing}
Laurens Van~der Maaten and Geoffrey Hinton.
\newblock Visualizing data using t-{SNE}.
\newblock \emph{Journal of Machine Learning Research}, 9\penalty0 (11), 2008.

\bibitem[Wang et~al.(2021)Wang, Ma, Zhu, and Yang]{Wang3dpoint}
Chunwei Wang, Chao Ma, Ming Zhu, and Xiaokang Yang.
\newblock Pointaugmenting: Cross-modal augmentation for 3d object detection.
\newblock In \emph{Computer Vision and Pattern Recognition}, pages 11794--11803, Virtual Event, 2021.

\bibitem[Wang et~al.(2022)Wang, Zhao, Peng, Zhu, Yang, Wang, Huang, Bilen, Wang, and You]{dd_CAFE}
Kai Wang, Bo Zhao, Xiangyu Peng, Zheng Zhu, Shuo Yang, Shuo Wang, Guan Huang, Hakan Bilen, Xinchao Wang, and Yang You.
\newblock Cafe: Learning to condense dataset by aligning features.
\newblock In \emph{Computer Vision and Pattern Recognition}, pages 12196--12205, New Orleans, LA, USA, 2022. {IEEE}.

\bibitem[Wang et~al.(2018)Wang, Zhu, Torralba, and Efros]{dd_begin}
Tongzhou Wang, Jun-Yan Zhu, Antonio Torralba, and Alexei~A Efros.
\newblock Dataset distillation.
\newblock \emph{arXiv preprint arXiv:1811.10959}, 2018.

\bibitem[Yin et~al.(2020)Yin, Molchanov, Alvarez, Li, Mallya, Hoiem, Jha, and Kautz]{deepinversion}
Hongxu Yin, Pavlo Molchanov, Jose~M Alvarez, Zhizhong Li, Arun Mallya, Derek Hoiem, Niraj~K Jha, and Jan Kautz.
\newblock Dreaming to distill: Data-free knowledge transfer via deepinversion.
\newblock In \emph{Computer Vision and Pattern Recognition}, pages 8715--8724, Virtual Event, 2020. {IEEE}.

\bibitem[Yin et~al.(2023)Yin, Xing, and Shen]{dd_sre2l}
Zeyuan Yin, Eric~P. Xing, and Zhiqiang Shen.
\newblock Squeeze, recover and relabel: Dataset condensation at imagenet scale from {A} new perspective.
\newblock In \emph{Neural Information Processing Systems}. NIPS, 2023.

\bibitem[You et~al.(2017)You, Xu, Xu, and Tao]{multi_teacher}
Shan You, Chang Xu, Chao Xu, and Dacheng Tao.
\newblock Learning from multiple teacher networks.
\newblock In \emph{International Conference on Knowledge Discovery and Data Mining}, pages 1285--1294, Halifax, NS, Canada, 2017. {ACM}.

\bibitem[Yu et~al.(2023)Yu, Liu, and Wang]{dd_comprehensive_review}
Ruonan Yu, Songhua Liu, and Xinchao Wang.
\newblock Dataset distillation: {A} comprehensive review.
\newblock \emph{arXiv preprint arXiv:2301.07014}, 2023.

\bibitem[Zagoruyko and Komodakis(2016)]{Zagoruyko2016WRN}
Sergey Zagoruyko and Nikos Komodakis.
\newblock Wide residual networks.
\newblock In \emph{British Machine Vision Conference}, York, UK, 2016. {BMVA} Press.

\bibitem[Zhang et~al.(2023)Zhang, Zhang, Lei, Mukherjee, Pan, Zhao, Ding, Li, and Xu]{dd_model_augmentation}
Lei Zhang, Jie Zhang, Bowen Lei, Subhabrata Mukherjee, Xiang Pan, Bo Zhao, Caiwen Ding, Yao Li, and Dongkuan Xu.
\newblock Accelerating dataset distillation via model augmentation.
\newblock In \emph{Computer Vision and Pattern Recognition}, Vancouver, BC, Canada, 2023. {IEEE}.

\bibitem[Zhang et~al.(2018)Zhang, Zhou, Lin, and Sun]{shufflenet}
Xiangyu Zhang, Xinyu Zhou, Mengxiao Lin, and Jian Sun.
\newblock Shufflenet: An extremely efficient convolutional neural network for mobile devices.
\newblock In \emph{Computer Vision and Pattern Recognition}, pages 6848--6856, 2018.

\bibitem[Zhao and Bilen(2021)]{dd_continual_learning_3}
Bo Zhao and Hakan Bilen.
\newblock Dataset condensation with differentiable siamese augmentation.
\newblock In \emph{International Conference on Machine Learning}, pages 12674--12685, Virtual Event, 2021. {PMLR}.

\bibitem[Zhao and Bilen(2023)]{dd_dist_matching}
Bo Zhao and Hakan Bilen.
\newblock Dataset condensation with distribution matching.
\newblock In \emph{Winter Conference on Applications of Computer Vision}, pages 6514--6523, Waikoloa, Hawaii, 2023. {IEEE}.

\bibitem[Zhao et~al.(2021)Zhao, Mopuri, and Bilen]{dd_gradient_matching}
Bo Zhao, Konda~Reddy Mopuri, and Hakan Bilen.
\newblock Dataset condensation with gradient matching.
\newblock In \emph{International Conference on Learning Representations}, Virtual Event, 2021. OpenReview.net.

\bibitem[Zhao et~al.(2022)Zhao, Cui, Song, Qiu, and Liang]{DKD}
Borui Zhao, Quan Cui, Renjie Song, Yiyu Qiu, and Jiajun Liang.
\newblock Decoupled knowledge distillation.
\newblock In \emph{Computer Vision and Pattern Recognition}, pages 11953--11962, 2022.

\bibitem[Zhou et~al.(2022)Zhou, Nezhadarya, and Ba]{dd_FRePo}
Yongchao Zhou, Ehsan Nezhadarya, and Jimmy Ba.
\newblock Dataset distillation using neural feature regression.
\newblock In \emph{Neural Information Processing Systems}, New Orleans, LA, USA, 2022. NIPS.

\end{thebibliography}
}
\clearpage
\setcounter{page}{1}
\onecolumn
\appendix
\section{Hyperparameter Settings}
\label{apd:hyperparameter_setting}\renewcommand{\arraystretch}{1.2}
\setlength{\tabcolsep}{0.78em}
\begin{table*}[!th]
\caption{G-VBSM hyperparameter settings on ImageNet-1k.}
\label{tab:hyperparameter_imagenet_1k}
\footnotesize
\centering
\resizebox{1.0\textwidth}{!}{
\begin{tabular}{c|cccccccc}
\toprule
\multicolumn{1}{c|}{\multirow{2}{*}{Phase}} & \multicolumn{1}{c}{\multirow{2}{*}{Optimizer}} & \multicolumn{1}{c}{\multirow{2}{*}{\makecell{Learning\\Rate}}}& \multicolumn{1}{c}{\multirow{2}{*}{\makecell{Optimizer\\Momentum}}} &
\multicolumn{1}{c}{\multirow{2}{*}{\makecell{Loss\\Function}}} & \multicolumn{1}{c}{\multirow{2}{*}{\makecell{Batch\\Size}}} &
\multicolumn{1}{c}{\multirow{2}{*}{\makecell{Epoch/\\Iteration}}} & \multicolumn{1}{c}{\multirow{2}{*}{Augmentation}} &
\multicolumn{1}{c}{\multirow{2}{*}{Others}}  \\
&&&&&&&\\ \hline
\specialrule{0em}{0.5pt}{0.5pt}
\makecell{Pre-trained model\\training} & SGD & 0.1 & 0.9 & cross-entropy & 256 & \makecell{Epoch\\100}  & RandomResizedCrop & - \\
\makecell{Data synthesis} & Adam & 0.1 & $\beta_1$,$\beta_2$=0.5,0.9 & \makecell{$\ell (f_\textrm{cand}(\tilde{X}),y) + \mathcal{L}^\prime_\textrm{BN}$\\$+ \mathcal{L}_\textrm{DD} + \mathcal{L}^\prime_\textrm{Conv}$} & 40 & \makecell{Iteration\\4000} & RandomResizedCrop & \makecell{$\beta_\textrm{dr}$=0.4, Backbone=\\$\{$ResNet18,MobileNetV2\\,EfficientNet-B0,ShuffleNetV2-0.5$\}$}\\
\makecell{Soft label\\generation} & - & - & - & - & 1024 & \makecell{Epoch\\300} & \makecell{RandomResizedCrop,\\CutMix} & \makecell{Backbone=\\$\{$ResNet18,MobileNetV2\\,EfficientNet-B0,ShuffleNetV2-0.5$\}$}\\
\makecell{Evaluation} & AdamW & 0.001 & $\beta_1$,$\beta_2$=0.9,0.999 & MSE+0.1$\times$GT & 1024 & \makecell{Epoch\\300} & \makecell{RandomResizedCrop,\\CutMix} & \makecell{Evaluation Model=\\\{ResNet18,ResNet50,ResNet101,\\MobileNetV2,Swin-Tiny,DeiT-Tiny\}} \\
\bottomrule
\end{tabular}}
\end{table*}\renewcommand{\arraystretch}{1.2}
\setlength{\tabcolsep}{0.78em}
\begin{table*}[!th]
\caption{G-VBSM hyperparameter settings on Tiny-ImageNet.}
\label{tab:hyperparameter_tiny_imagenet}
\footnotesize
\centering
\resizebox{1.0\textwidth}{!}{
\begin{tabular}{c|cccccccc}
\toprule
\multicolumn{1}{c|}{\multirow{2}{*}{Phase}} & \multicolumn{1}{c}{\multirow{2}{*}{Optimizer}} & \multicolumn{1}{c}{\multirow{2}{*}{\makecell{Learning\\Rate}}}& \multicolumn{1}{c}{\multirow{2}{*}{\makecell{Optimizer\\Momentum}}} &
\multicolumn{1}{c}{\multirow{2}{*}{\makecell{Loss\\Function}}} & \multicolumn{1}{c}{\multirow{2}{*}{\makecell{Batch\\Size}}} &
\multicolumn{1}{c}{\multirow{2}{*}{\makecell{Epoch/\\Iteration}}} & \multicolumn{1}{c}{\multirow{2}{*}{Augmentation}} &
\multicolumn{1}{c}{\multirow{2}{*}{Others}}  \\
&&&&&&&\\ \hline
\specialrule{0em}{0.5pt}{0.5pt}
\makecell{Pre-trained model\\training} & SGD & 0.1 & 0.9 & cross-entropy & 128 & \makecell{Epoch\\50}  &  \makecell{RandomCrop\\RandomHorizontalFlip}  & - \\
\makecell{Data synthesis} & Adam & 0.05 & $\beta_1$,$\beta_2$=0.5,0.9 & \makecell{$\ell (f_\textrm{cand}(\tilde{X}),y) + \mathcal{L}^\prime_\textrm{BN}$\\$+ \mathcal{L}_\textrm{DD} + \mathcal{L}^\prime_\textrm{Conv}$} & 50 & \makecell{Iteration\\4000} & RandomResizedCrop & \makecell{$\beta_\textrm{dr}$=0.4, Backbone=\\$\{$ResNet18,128-width ConvNet,MobileNetV2\\,WRN-16-2,ShuffleNetV2-0.5$\}$} \\
\makecell{Soft label\\generation} & - & - & - & - & 128 & \makecell{Epoch\\100} & \makecell{RandomResizedCrop,\\CutMix} & \makecell{Backbone=\\$\{$ResNet18,128-width ConvNet,MobileNetV2\\,WRN-16-2,ShuffleNetV2-0.5$\}$}\\
\makecell{Evaluation} & SGD & \makecell{0.2, 0.1 and 0.1 on \\ ResNet18, ResNet50 and ResNet101\\, respectively} & 0.9 & MSE+0.1$\times$GT & 128 & \makecell{Epoch\\100} & \makecell{RandomResizedCrop,\\CutMix} & \makecell{Evaluation Model=\\\{ResNet18,ResNet50,ResNet101\}} \\
\bottomrule
\end{tabular}}
\end{table*}\renewcommand{\arraystretch}{1.2}
\setlength{\tabcolsep}{0.78em}
\begin{table*}[!th]
\caption{G-VBSM hyperparameter settings on CIFAR-10/100.}
\label{tab:hyperparameter_cifar_100}
\footnotesize
\centering
\resizebox{1.0\textwidth}{!}{
\begin{tabular}{c|cccccccc}
\toprule
\multicolumn{1}{c|}{\multirow{2}{*}{Phase}} & \multicolumn{1}{c}{\multirow{2}{*}{Optimizer}} & \multicolumn{1}{c}{\multirow{2}{*}{\makecell{Learning\\Rate}}}& \multicolumn{1}{c}{\multirow{2}{*}{\makecell{Optimizer\\Momentum}}} &
\multicolumn{1}{c}{\multirow{2}{*}{\makecell{Loss\\Function}}} & \multicolumn{1}{c}{\multirow{2}{*}{\makecell{Batch\\Size}}} &
\multicolumn{1}{c}{\multirow{2}{*}{\makecell{Epoch/\\Iteration}}} & \multicolumn{1}{c}{\multirow{2}{*}{Augmentation}} &
\multicolumn{1}{c}{\multirow{2}{*}{Others}}  \\
&&&&&&&\\ \hline
\specialrule{0em}{0.5pt}{0.5pt}
\makecell{Pre-trained model\\training} & SGD & 0.05 & 0.9 & cross-entropy & 64 & \makecell{Epoch\\50 and 5 on \\CIFAR-100 and CIFAR-10\\, respectively}  & \makecell{RandomCrop\\RandomHorizontalFlip} & - \\
\makecell{Data synthesis} & Adam & 0.05 & $\beta_1$,$\beta_2$=0.5,0.9 & \makecell{$\ell (f_\textrm{cand}(\tilde{X}),y) + \mathcal{L}^\prime_\textrm{BN}$\\$+ \mathcal{L}_\textrm{DD} + \mathcal{L}^\prime_\textrm{Conv}$} & 50 & \makecell{Iteration\\4000} & RandomResizedCrop & \makecell{$\beta_\textrm{dr}$=0.4, Backbone=\\$\{$ResNet18,128-width ConvNet,MobileNetV2\\,WRN-16-2,ShuffleNetV2-0.5$\}$}\\
\makecell{Soft label\\generation} & - & - & - & - & \makecell{64 or $|\mathcal{S}|$\\($|\mathcal{S}|\leq$100)} & \makecell{Epoch\\1000} & \makecell{RandomResizedCrop,\\CutMix} & \makecell{Backbone=\\$\{$ResNet18,128-width ConvNet,MobileNetV2\\,WRN-16-2,ShuffleNetV2-0.5$\}$}\\
\makecell{Evaluation} & \makecell{SGD and AdamW on \\CIFAR-100 and CIFAR-10\\, respectively} &  \makecell{0.1 and 0.001 on \\CIFAR-100 and CIFAR-10\\, respectively} &  \makecell{0.9 and $\beta_1$,$\beta_2$=0.9,0.999 on \\CIFAR-100 and CIFAR-10\\, respectively} & MSE+0.15$\times$GT & \makecell{64 or $|\mathcal{S}|$\\($|\mathcal{S}|\leq$100)} & \makecell{Epoch\\1000} & \makecell{RandomResizedCrop,\\CutMix} & \makecell{Evaluation Model=\\\{128-width ConvNet,ResNet18\}} \\
\bottomrule
\end{tabular}}
\end{table*}
\noindent Here, we present the hyperparameter settings of G-VBSM in the pre-trained model training (\textit{i.e.}, Squeeze in SRe2L), the data synthesis (\textit{i.e.}, Recover in SRe2L), the soft label generation (\textit{i.e.}, Relabel in SRe2L), and the evaluation phases in Tables~\ref{tab:hyperparameter_imagenet_1k} (ImageNet-1k),~\ref{tab:hyperparameter_tiny_imagenet} (Tiny-ImageNet), and~\ref{tab:hyperparameter_cifar_100} (CIFAR-10/CIFAR-100).
The hyperparameter settings for the ImageNet-1k and Tiny-ImageNet datasets predominantly adhere to SRe2L~\cite{dd_sre2l}. Furthermore, the settings for CIFAR-10/CIFAR-100 draw upon the classical knowledge distillation framework~\cite{vanillakd,DIST,DKD,CRD}. Notably, we employ the same evaluation model (\textit{i.e.} 128-width ConvNet) and identical number of epochs (\textit{i.e.} 1000) during the evaluation phase on CIFAR-10/CIFAR-100 as those used in the prior dataset distillation approaches~\cite{dd_datadam,dd_mtt}, ensuring experimental fairness.

\paragraph{The Consistency of Backbone used in Data Synthesis and Soft Label Generation.} In all experiments conducted on different datasets, we maintain the same architectures and identical parameters of the pre-trained model for data synthesis and soft label generation. Similar to SRe2L, our exploratory studies revealed that preserving the consistency of the backbone results in the best generalization ability for the distilled dataset.

\paragraph{The Hyperparameter $\beta_\textrm{dr}$.} Given G-VBSM's computational efficiency on ImageNet-1k under IPC 10, which serves as the benchmark for the majority of our ablation studies, we set $\beta_\textrm{dr}$ to 0.0 for this specific benchmark. For the remaining experiments, including Tiny-ImageNet, CIFAR-10 and CIFAR-100, $\beta_\textrm{dr}$ is set to 0.4.

\paragraph{The Weights of the Loss Function.} To underscore the generalization and applicability of our proposed G-VBSM, we intentionally avoid setting the weights of any loss functions, except for MSE+$\gamma$$\times$GT, in a bespoke manner. To be specific, the weights for both $\mathcal{L}^\prime_\textrm{BN}$ and $\mathcal{L}^\prime_\textrm{Conv}$ are established at 0.01 for ImageNet-1k, consistent with the weight of $\mathcal{L}_\textrm{BN}$, which SRe2L is set as 0.01 for ImageNet-1k. Since we transposed the loop (\textit{i.e.}, translate the original loop to the reorder loop), the weights for $\mathcal{L}^\prime_\textrm{BN}$ and $\mathcal{L}^\prime_\textrm{Conv}$ are set at 0.01 for Tiny-ImageNet, different from SRe2L, which assigns a weight of 1.0 to $\mathcal{L}_\textrm{BN}$ for the same dataset. Due to SRe2L was not evaluated on CIFAR-10 and CIFAR-100, we empirically adjusted the weights for $\mathcal{L}^\prime_\textrm{BN}$ and $\mathcal{L}^\prime_\textrm{Conv}$ to 0.01 in our experiments. Additionally, the weight of $\mathcal{L}_\textrm{DD}$ is consistently applied at 1.0 across all datasets. Through empirical validation in our experiments, we establish that the performance of the distilled dataset \mbox{--} when the weight of $\mathcal{L}_\textrm{DD}$ is configured as \{0.1, 1.0, 10.0\} \mbox{--} remains precisely identical.

\section{The Derivation of ``match in the form of score distillation sampling''}
\label{apd:derivation_bn}

Our proposed novel loss function, denoted as $\mathcal{L}_{\textrm{BN}}^{\prime}(\tilde{X})$, draws inspiration from score distillation sampling (\textbf{SDS})~\cite{iclr2023_dreamfusion}. It is employed to mitigate the performance degradation that arises from directly substituting the original loop with the reorder loop. Here, we give the detailed derivation of $\mathcal{L}_{\textrm{BN}}^{\prime}(\tilde{X})$ to facilitate understanding. First, we give the gradient of the original loss function $\mathcal{L}_{\textrm{BN}}(\tilde{X})$ with respect to $\tilde{X}$:

\begin{equation}
\small
\begin{aligned}
\frac{\partial \mathcal{L}_{\textrm{BN}}(\tilde{X})}{\partial \tilde{X}} &= \sum_l \frac{\partial \mu_l(\tilde{X})}{\partial \tilde{X}} \frac{\mu_l(\tilde{X}) - {\textbf{\textrm{BN}}_l^\textrm{CM}}}{||\mu_l(\tilde{X}) - {\textbf{\textrm{BN}}_l^\textrm{CM}}||_2} +\frac{\partial \sigma^2_l(\tilde{X})}{\partial \tilde{X}}\frac{\sigma^2_l(\tilde{X}) - {\textbf{\textrm{BN}}_l^\textrm{CV}}}{||\sigma^2_l(\tilde{X}) - {\textbf{\textrm{BN}}_l^\textrm{CV}}||_2}. \\
\end{aligned}
\label{eq:appendx_b_1}
\end{equation}
In Eq.~\ref{eq:appendx_b_1}, $\frac{\mu_l(\tilde{X}) - {\textbf{\textrm{BN}}_l^\textrm{CM}}}{||\mu_l(\tilde{X}) - {\textbf{\textrm{BN}}_l^\textrm{CM}}||_2}$ and $\frac{\sigma^2_l(\tilde{X}) - {\textbf{\textrm{BN}}_l^\textrm{CV}}}{||\sigma^2_l(\tilde{X}) - {\textbf{\textrm{BN}}_l^\textrm{CV}}||_2}$ are unit vectors that dominate the direction of the gradient descent in the data synthesis process. As analyzed in Sec.\ref{sec:method_dd}, the precise global statistics generated by all past batches are feasible to assist in matching between the limited statistics generated by the current batch and ${\textbf{\textrm{BN}}_l^\textrm{CM}}$ as well as ${\textbf{\textrm{BN}}_l^\textrm{CV}}$. We utilize EMA to update the statistics $\mu_l^\textrm{total} $ and $\sigma^{2,\textrm{total}}_l$ generated by all past batches:
\begin{equation}
\small
\begin{aligned}
& \mu_l^\textrm{total} = \alpha \mu_l^\textrm{total} + (1 - \alpha) \mu_l(\tilde{X}),\sigma^{2,\textrm{total}}_l= \alpha \sigma^{2,\textrm{total}}_l + (1 - \alpha) \sigma^2_l(\tilde{X}).\\
\end{aligned}
\label{eq:appendx_b_2}
\end{equation}
We can achieve the SDS-like loss by simply replacing $\frac{\mu_l(\tilde{X}) - {\textbf{\textrm{BN}}_l^\textrm{CM}}}{||\mu_l(\tilde{X}) - {\textbf{\textrm{BN}}_l^\textrm{CM}}||_2}$ with $\frac{\mu^\textrm{total}_l(\tilde{X}) - {\textbf{\textrm{BN}}_l^\textrm{CM}}}{||\mu^\textrm{total}_l(\tilde{X}) - {\textbf{\textrm{BN}}_l^\textrm{CM}}||_2}$ and $\frac{\sigma^2_l(\tilde{X}) - {\textbf{\textrm{BN}}_l^\textrm{CV}}}{||\sigma^2_l(\tilde{X}) - {\textbf{\textrm{BN}}_l^\textrm{CV}}||_2}$ with $\frac{\sigma^{2,\textrm{total}}_l(\tilde{X}) - {\textbf{\textrm{BN}}_l^\textrm{CV}}}{||\sigma^{2,\textrm{total}}_l(\tilde{X}) - {\textbf{\textrm{BN}}_l^\textrm{CV}}||_2}$. In this way, the direction of gradient descent for data synthesis is no longer determined by the imprecise statistics of the single current batch, which ultimately improves the quality of the synthetic data and its ability to generalize to unseen evaluation models. In practice, we can implement the replacement easily with Pytorch's~\cite{pytorch} $\textrm{stop\_grad}(\cdot)$ operator:
\begin{equation}
\small
\begin{aligned}
& \mathcal{L}_{\textrm{BN}}^{\prime}(\tilde{X}) = \sum_l || \mu_l(\tilde{X}) - {\textbf{\textrm{BN}}_l^\textrm{CM}} - \textrm{stop\_grad}(\mu_l(\tilde{X}) - \mu_l^\textrm{total})||_2 +|| \sigma^2_l(\tilde{X}) - {\textbf{\textrm{BN}}_l^\textrm{CV}} - \textrm{stop\_grad}(\sigma^2_l(\tilde{X}) - \sigma^{2,\textrm{total}}_l)||_2. \\
\end{aligned}
\label{eq:appendx_b_3}
\end{equation}
We can find the gradient of $\mathcal{L}_{\textrm{BN}}^{\prime}(\tilde{X})$ with respect to $\tilde{X}$ by derivation as
\begin{equation}
\small
\begin{aligned}
\frac{\partial \mathcal{L}_{\textrm{BN}}^{\prime}(\tilde{X})}{\partial \tilde{X}} &= \sum_l \frac{\partial \mu_l(\tilde{X})}{\partial \tilde{X}} \frac{\mu^\textrm{total}_l(\tilde{X}) - {\textbf{\textrm{BN}}_l^\textrm{CM}}}{||\mu^\textrm{total}_l(\tilde{X}) - {\textbf{\textrm{BN}}_l^\textrm{CM}}||_2} +\frac{\partial \sigma^2_l(\tilde{X})}{\partial \tilde{X}}\frac{\sigma^{2,\textrm{total}}_l(\tilde{X}) - {\textbf{\textrm{BN}}_l^\textrm{CV}}}{||\sigma^{2,\textrm{total}}_l(\tilde{X}) - {\textbf{\textrm{BN}}_l^\textrm{CV}}||_2}. \\
\end{aligned}
\label{eq:appendx_b_4}
\end{equation}
Clearly, $\mathcal{L}_{\textrm{BN}}^{\prime}(\tilde{X})$ effectively achieves our primary purpose. Additionally, our ablation studies in Sec.~\ref{sec:ablation_studies} empirically demonstrate that the ``match in the form of score distillation sampling'' strategy is remarkable.

\section{Additional Ablation Experiments on ImageNet-1k}
\label{apd:imagenet_add}
\begin{table*}[!h]
\centering
\begin{tabular}{cc|cccc}\toprule
 $\mathcal{L}^\prime_\textrm{BN}$ & $\mathcal{L}^\prime_\textrm{Conv}$ & ResNet18 & ResNet50 & ResNet101\\\hline
\checkmark & & 27.8\% & 33.4\% & 35.5\% \\
&\checkmark & 24.0\% & 26.1\% & 30.4\% \\
\checkmark &\checkmark &  \cellcolor{gray!25} 31.4\% &  \cellcolor{gray!25} 35.4\% &  \cellcolor{gray!25} 38.2\% \\
\bottomrule
\end{tabular}
\caption{Ablation study about $\mathcal{L}^\prime_\textrm{BN}$ and $\mathcal{L}^\prime_\textrm{Conv}$ in the synthetic data phase on ImageNet-1k. Meanwhile, ResNet \{18, 50, 101\} represent evaluation models.}
\label{tab:imagenet_bn_conv}
\vspace{-0.1in}
\end{table*}

\noindent This section presents ablation experiments for $\mathcal{L}^\prime_\textrm{BN}$ and $\mathcal{L}^\prime_\textrm{Conv}$ to underscore their equal importance. As illustrated in Table~\ref{tab:imagenet_bn_conv}, omitting either $\mathcal{L}^\prime_\textrm{BN}$ or $\mathcal{L}^\prime_\textrm{Conv}$ from the entire loss function during data synthesis phase leads to a performance decline. Hence, conducting the ``local-match-global'' matching via both $\mathcal{L}^\prime_\textrm{BN}$ and $\mathcal{L}^\prime_\textrm{Conv}$ is essential.

\section{Exploratory Studies on CIFAR-10/100}
\label{apd:cifar_10_and_100_add}
\paragraph{The Choice of Candidate Backbones in GBM.} Under IPC 10 on CIFAR-100, we evaluated candidate backbones \{ResNet18, MobileNetV2, WRN-16-2, ShuffleNetV2-0.5\}, omitting the 128-width ConvNet, during the data synthesis and soft label generation phases. In addition, we kept other hyperparameters consistent as shown in Table~\ref{tab:hyperparameter_cifar_100} and obtained the 128-width ConvNet evaluation performance as 32.8\%. However, incorporating the 128-width ConvNet into the candidate backbones increased the accuracy from 32.8\% to 38.7\%. It's important to mention that the 128-width ConvNet solely utilizes GroupNorm, not BatchNorm. This enhancement to 38.7\% was accomplished by relying solely on statistics within Convolution, substantiating that statistics in BatchNorm may not be the only option in the data synthesis phase.

\paragraph{The Number of Epochs in the Pre-Trained Model Training Phase.} \begin{table*}[!h]
\centering
\begin{tabular}{c|cccc}\toprule
Evaluation Model\verb|\|Epoch & 5 & 10 & 20 & 40 \\\hline
128-width ConvNet & \cellcolor{gray!25} 46.5\% & 45.8\% & 42.5\% & 42.1\% \\
\bottomrule
\end{tabular}
\caption{Ablation study about the number of epochs in pre-trained model training phase. We maintain the consistency of other hyperparameters as presented in Table~\ref{tab:hyperparameter_cifar_100}.}
\label{tab:c10_pre_train_epoch}
\vspace{-0.1in}
\end{table*}

 As illustrated in Table~\ref{tab:c10_pre_train_epoch}, fewer pre-training epochs on CIFAR-10 enhance the generalization of the distilled dataset. This finding could provide an explanation for the remarkable performance achieved by traditional algorithms~\cite{dd_mtt,dd_begin} on CIFAR-10, even when they employ models with random initializations.  As a result, this ablation study informed our decision to pre-train models on CIFAR-10 for only 5 epochs. More important, as our experiments transition from ImageNet-1k to Tiny-ImageNet to CIFAR-100, and finally to CIFAR-10, the dataset complexity reduces, and the ideal number of pre-training epochs successively decreases from 100 to 50, to 50, and finally to 5. The most intuitive and empirical extrapolation is due to the complexity of the dataset, and we believe that this conclusion may be of some inspiration to other researchers.

\section{Additional Explanation of Data Densification}
\label{sec:add_dd_explan}
Here we provide theoretical proofs within Eq.~\ref{eq:cos_tau} to show that entropy $H(\Sigma_y/\tau)$ ($\tau>1$) is greater than $H(\Sigma_y)$, thus increasing $H(\Sigma_y)$ through Eq.~\ref{eq:dd_2}, which ultimately improves the entropy of the eigenvalues and ensures the diversity of data.
\begin{equation}
\begin{aligned}
         & H(z/\tau) - H(z) = (\tau-1)\frac{\partial H}{\partial \tau}(\tau^\prime),\ s.t.\ 1\leq \tau^\prime \leq \tau, \textrm{define}\ z=\Sigma_y \textrm{for}\ \textrm{convenience}\\
         & = -\frac{\tau-1}{\tau^\prime}\sum_i [(\log(\frac{e^{z_i/\tau^\prime}}{\sum_j e^{z_j/\tau^\prime}})+1)(\frac{e^{z_i/\tau^\prime}}{\sum_j e^{z_j/\tau^\prime}})\\
         &(\frac{-z_i/\tau^\prime\sum_j (e^{z_j/\tau^\prime})+\sum_j(e^{z_j/\tau^\prime}z_j/\tau^\prime)}{\sum_{j}e^{z_j/\tau^\prime}})] = -\frac{\tau-1}{\tau^\prime}[\sum_i \log(\frac{e^{z_i/\tau^\prime}}{\sum_j e^{z_j/\tau^\prime}}) \\
         & (\frac{e^{z_i/\tau^\prime}}{\sum_j e^{z_j/\tau^\prime}})(\frac{-z_i/\tau^\prime\sum_j (e^{z_j/\tau^\prime})+\sum_j(e^{z_j/\tau^\prime}z_j/\tau^\prime)}{\sum_{j}e^{z_j/\tau^\prime}})- \blue{\bcancel{\frac{\sum_j z_j/\tau^\prime e^{z_j/\tau^\prime}}{\sum_j e^{z_j/\tau^\prime}}}}\\
         & +\blue{\bcancel{\frac{\sum_j z_j/\tau^\prime e^{z_j/\tau^\prime}}{\sum_j e^{z_j/\tau^\prime}}}}] = -\frac{\tau-1}{\tau^\prime}\sum_j (\frac{e^{z_i/\tau^\prime}}{\sum_j e^{z_j/\tau^\prime}})\log(\frac{e^{z_i/\tau^\prime}}{\sum_j e^{z_j/\tau^\prime}})\\
         & [-z_i/\tau^\prime + \frac{\sum_j e^{z_j/\tau^\prime}z_j/\tau^\prime}{\sum_j e^{z_j/\tau^\prime}}] = \frac{\tau-1}{\tau^\prime}[\sum_i (z_i/\tau^\prime)^2e^{z_i/\tau^\prime}- \sum_i (z_i/\tau^\prime e^{z_i/\tau^\prime})\\
         & \frac{\sum_i (z_i/\tau^\prime e^{z_i/\tau^\prime})}{\sum_i (e^{z_i/\tau^\prime})} +\log(\sum_i e^{z_i/\tau^\prime})(\blue{\bcancel{\sum_i z_i/\tau^\prime e^{z_i/\tau^\prime}}}-\blue{\bcancel{\sum_i z_i/\tau^\prime e^{z_i/\tau^\prime})}}] >0. \\ 
    \end{aligned}
    \label{eq:cos_tau}
\end{equation}

\section{Statistics Visualization}\begin{figure}[!h]
\centering
\includegraphics[width=0.8\textwidth]{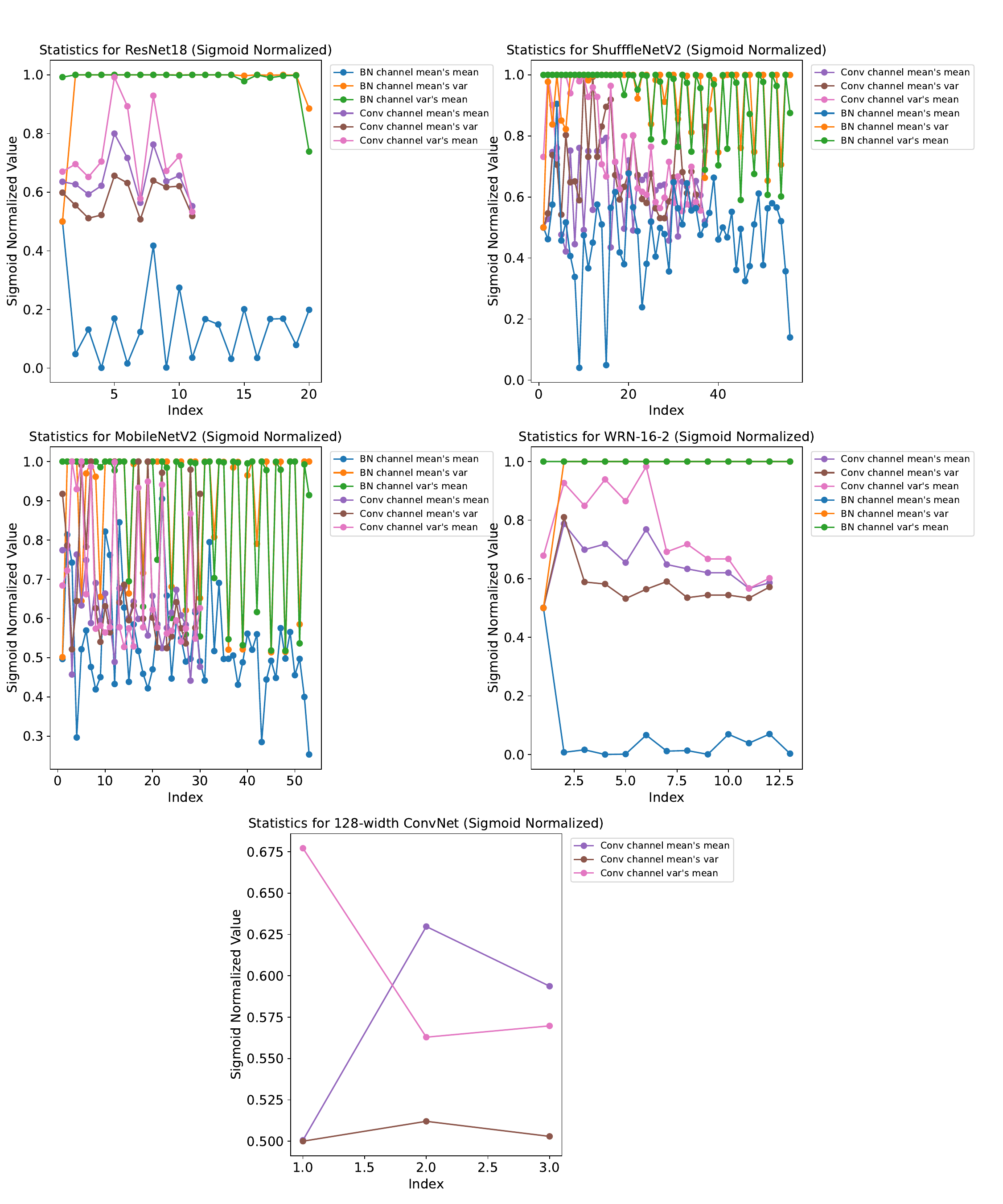}
        \caption{Statistics visualization across various backbones \{ResNet18, ShuffleNetV2, MobileNetV2, WRN-16-2, 128-width ConvNet\} on CIFAR-100.}
        \label{fig:network_statistics_plots}
\end{figure}
\noindent It is important to emphasize that Convolution and BatchNorm offer different supervisory information in statistics. As a result, G-VBSM is more effective than SRe2L when optimized statistics in Convolution and BatchNorm together. For clarity, we visualized the statistics in the pre-trained models \{ResNet18, 128-width ConvNet, MobileNetV2, WRN-16-2, ShuffleNetV2-0.5\} on CIFAR-100 in Fig.~\ref{fig:network_statistics_plots}. In each subplot of Fig.~\ref{fig:network_statistics_plots}, the horizontal axis denotes the layer index (with orthogonal indexes for Convolution and BatchNorm), while the vertical axis shows the post-sigmoid normalized result. Due to the extensive dimensions of channel mean and channel variance, we calculate only their mean and variance for visualization. Furthermore, since BatchNorm is not included in 128-width ConvNet, only Convolution statistics are visualized. From Fig.~\ref{fig:network_statistics_plots}, we can conclude that the values of the statistics in Convolution and BatchNorm are different in any model, which indicates that G-VBSM is significant and can enhance the generalization of the distilled dataset as demonstrated in Fig.~\ref{fig:efficiency_vs_effectiveness}.
\begin{figure}[!h]
\begin{center}
\includegraphics[width=0.65\linewidth,trim={0cm 0cm 0cm 0cm},clip]{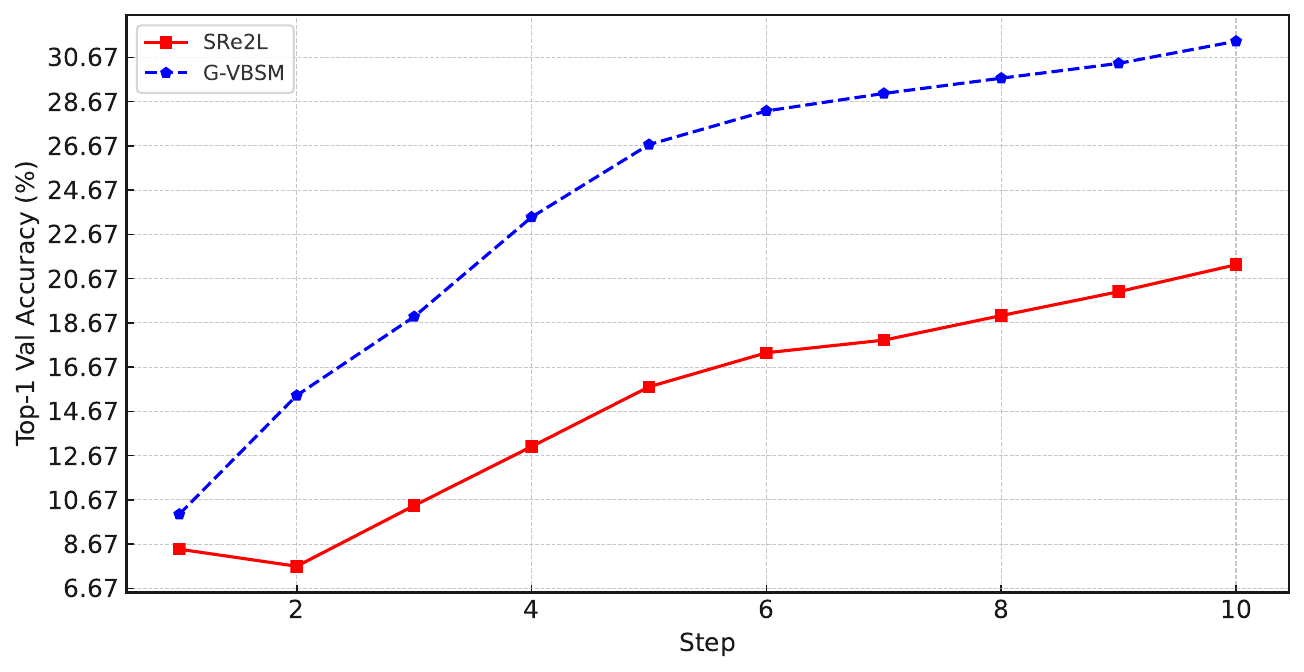}
\vspace{-15pt}
\end{center}
   \caption{The experimental result of continual learning application on ImageNet-1k under IPC 10.}
\label{fig:cl_imagenet_1k}
\vspace{-12pt}
\end{figure}
\section{Continual Learning Application}
Many data condensation algorithms~\cite{dd_dist_matching,dd_sre2l,dd_model_augmentation} have evaluated the generalization ability of distilled datasets in continual learning. We follow the class-incremental learning approach\footnote{This involves gradually increasing the number of classes and combining previously stored data with newly acquired data to train a model from scratch.} adopted in DM~\cite{dd_dist_matching} for performing this task. Similar to the ablation studies of the main paper, our experiments are conducted on the full 224$\times$224 ImageNet-1k, underscoring that G-VBSM is intended for use with large-scale datasets. We conduct class incremental learning with ResNet18 on the 10-step class-incremental learning strategy under 10 IPC. The experimental results are illustrated in Fig.~\ref{fig:cl_imagenet_1k}. We can discover that G-VBSM significantly outperforms SRe2L, thus confirming the usefulness and effectiveness of G-VBSM.

\section{Date Free Pruning Application}
\begin{table}[!h]
\centering
\renewcommand\arraystretch{0.95}
\resizebox{0.6\textwidth}{!}{%
\begin{tabular}{c|cccc}
\toprule
\multirow{2}{*}{\makecell{Data Free Pruning\\(ImageNet-1k, VGG-A, 50\% Pruned)}} & IPC 10 & IPC 10 & IPC 50 & IPC 50 \\
& SRe2L & SRe2L+DD & SRe2L & SRe2L+DD \\\hline
Top-1 Val Accuracy & 12.5\% & \cellcolor{gray!25} 12.9\% & 31.7\% & \cellcolor{gray!25} 32.8\% \\\bottomrule
\end{tabular}}
\caption{The experimental result of data free pruning application on IamgeNet-1k.}
\label{fig:data_free_pruning}
\end{table}
{\em Data Free Pruning of Slimming} aims to reduce the model size and decrease the run-time memory footprint simultaneously for convolutional nerual network. We argue that the distilled dataset facilitates efficient data-free pruning. To substantiate this claim, we conduct experiments on ImageNet-1k with IPC 10. As illustrated in Table~\ref{fig:data_free_pruning}, data densification enhances downstream knowledge transfer as above by increasing synthesized data diversity and significantly boosting SRe2L.

\section{Synthetic Data Visualization}
\label{apd:visualization}
\begin{figure}[!t]
\centering
\includegraphics[width=\textwidth]{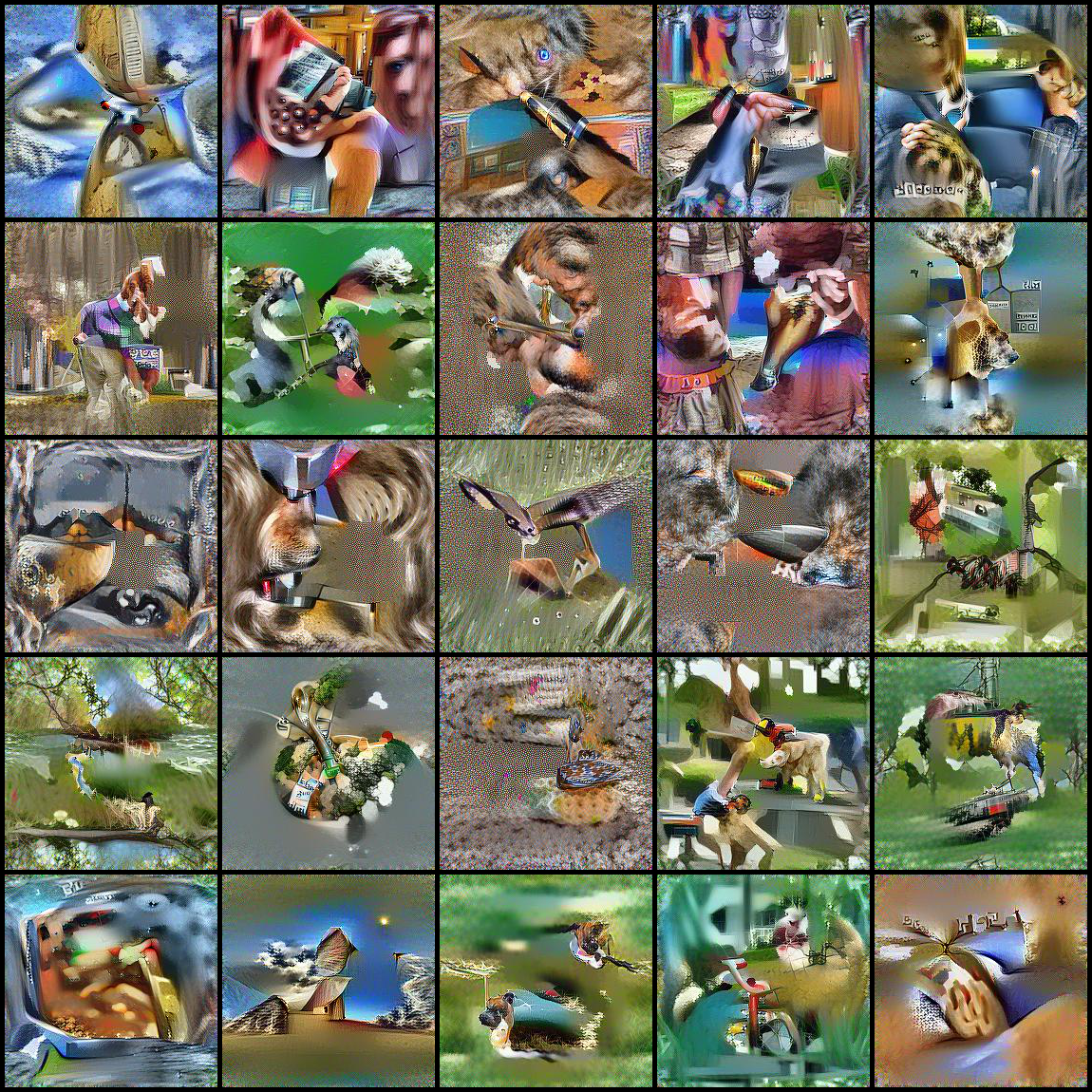}
        \caption{Synthetic data visualization on ImageNet-1k randomly selected from G-VBSM.}
        \label{fig:vis_imagenet_1k}
\end{figure}

\begin{figure}[!t]
\centering
\includegraphics[width=\textwidth]{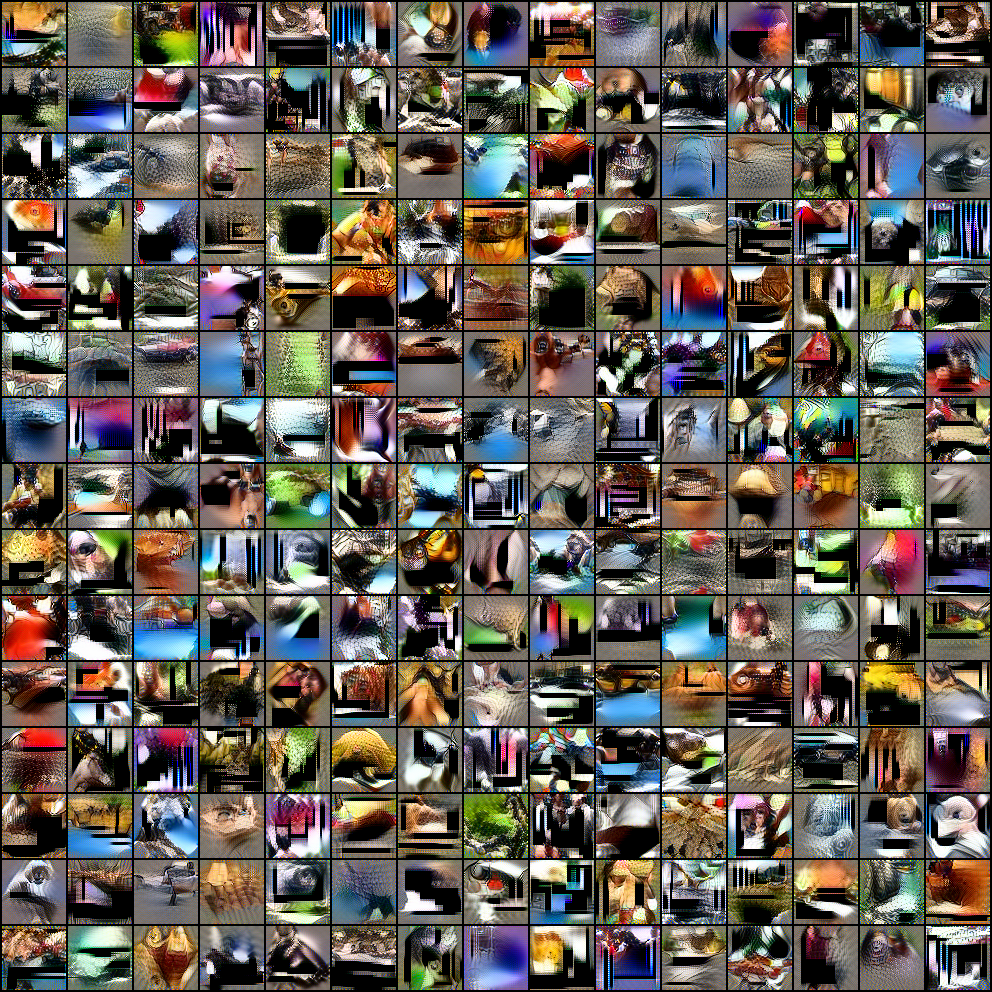}
        \caption{Synthetic data visualization on Tiny-ImageNet randomly selected from G-VBSM.}
        \label{fig:vis_tiny_imagenet}
\end{figure}

\begin{figure}[!t]
\centering
\includegraphics[width=\textwidth]{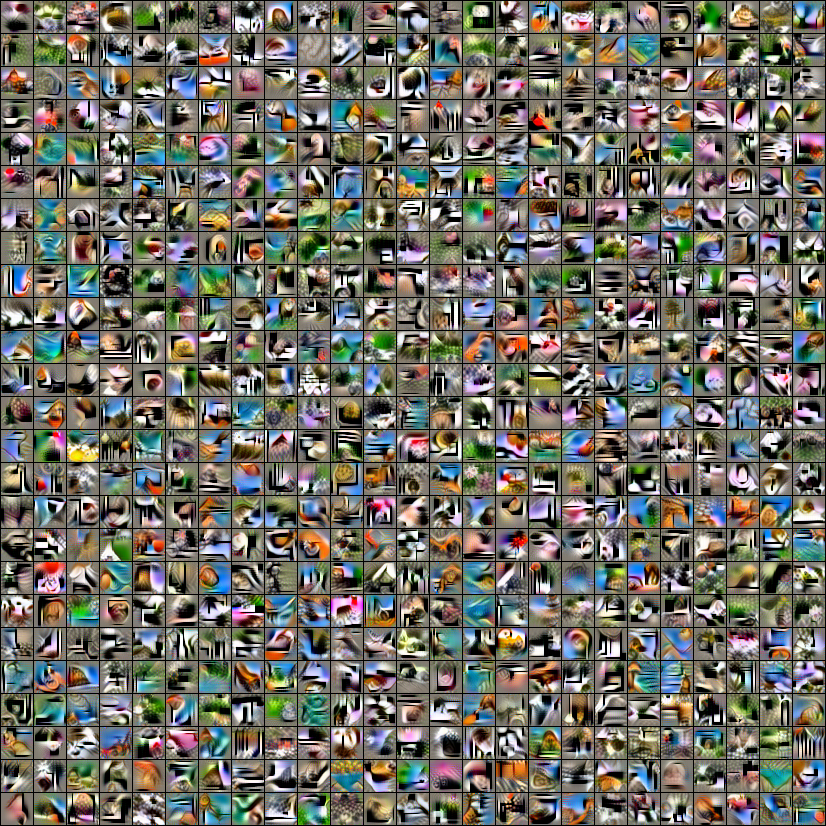}
        \caption{Synthetic data visualization on CIFAR-100 randomly selected from G-VBSM.}
        \label{fig:vis_cifar_100}
\end{figure}

\begin{figure}[!t]
\centering
\includegraphics[width=\textwidth]{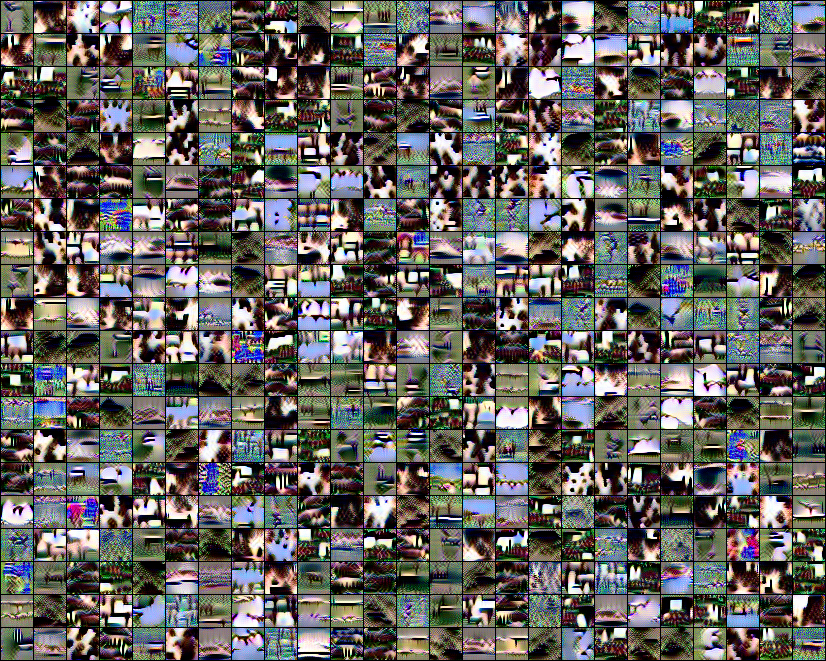}
        \caption{Synthetic data visualization on CIFAR-10 randomly selected from G-VBSM.}
        \label{fig:vis_cifar_10}
\end{figure}

We provide more visualization results on synthetic data randomly selected from G-VBSM in Figs.~\ref{fig:vis_imagenet_1k} (ImageNet-1k),~\ref{fig:vis_tiny_imagenet} (Tiny-ImageNet),~\ref{fig:vis_cifar_100} (CIFAR-100) and~\ref{fig:vis_cifar_10} (CIFAR-10).

\clearpage

\end{document}